\documentclass[11pt]{article}

\usepackage[final]{acl}

\usepackage{times}
\usepackage{latexsym}

\usepackage[T1]{fontenc}
\usepackage[utf8]{inputenc}

\usepackage{microtype}

\usepackage{inconsolata}

\usepackage{graphicx}
\usepackage{amsmath}
\usepackage{amsthm}
\usepackage{booktabs}
\usepackage{pifont}
\usepackage[table]{xcolor}
\usepackage{tabularx}
\usepackage{array}
\usepackage{multirow}

\definecolor{lbgreen}{RGB}{226, 242, 230}
\definecolor{lbred}{RGB}{252, 232, 230}
\definecolor{lbgray}{RGB}{240, 240, 240}
\definecolor{lbblue}{RGB}{231, 239, 252}

\newcommand{\cmarkcol}{\textcolor{green!60!black}{\ding{51}}}
\newcommand{\xmarkcol}{\textcolor{red!75!black}{\ding{55}}}
\newcommand{\cmark}{\ding{51}}
\newcommand{\xmark}{\ding{55}}
\title{LongBEL: Long-Context and Document-Consistent Biomedical Entity Linking}

\author{
 \textbf{Adam Remaki\textsuperscript{1}},
 \textbf{Xavier Tannier\textsuperscript{1}},
 \textbf{Christel Gérardin\textsuperscript{1,2}}
\\
\textsuperscript{1}\fontsize{10.5pt}{11pt}\selectfont Sorbonne Université, Inserm, Université Sorbonne Paris Nord, Limics, 75006 Paris, France,\\
\textsuperscript{2}\fontsize{10.5pt}{11pt}\selectfont Service de médecine interne, Hôpital Tenon, Assistance Publique - Hôpitaux de Paris, Paris, France
\\
 \small{
   \textbf{Correspondence:} \href{mailto:adam.remaki@etu.sorbonne-universite.fr}{adam.remaki@etu.sorbonne-universite.fr}
 }
}

\begin{document}
\maketitle
\begin{abstract}
Biomedical entity linking maps textual mentions to concepts in structured knowledge bases such as UMLS or SNOMED CT. Most generative BEL systems link each mention independently, using only the mention or its surrounding sentence. This ignores dependencies between mentions in the same document and can lead to inconsistent predictions, especially when the same concept appears under different surface forms. We introduce \textbf{LongBEL}, a document-level generative framework that combines full-document context with a memory of previous predictions. To make this memory robust, LongBEL is trained with cross-validated predictions rather than gold labels, reducing the mismatch between training and inference and limiting cascading errors. Experiments on five biomedical benchmarks across English, French, and Spanish show that LongBEL improves over sentence-level generative baselines, with the largest gains on datasets where concepts frequently recur within documents. An ensemble of local, global, and memory-based variants achieves the best results across all benchmarks. Further analysis shows that the largest gains occur on recurring concepts, suggesting that LongBEL mainly improves document-level consistency rather than isolated mention disambiguation.
\end{abstract}

\section{Introduction}

Biomedical Entity Linking (BEL) maps textual mentions in biomedical documents to concepts in structured knowledge bases (KBs) such as the Unified Medical Language System (UMLS) or SNOMED CT~\cite{bodenreider_unified_2004,donnelly_snomed-ct_2006}. This normalization step is essential for biomedical information extraction, as it resolves lexical variation and allows mentions referring to the same concept to be aggregated~\cite{krauthammer_term_2004}.

Recent neural models have substantially improved BEL performance, especially through contextualized representations. However, most state-of-the-art BEL systems still link mentions independently, using only local context~\cite{chen_comprehensive_2026}. This is limiting in biomedical documents, where concepts can reappear through abbreviations, synonyms, or partial expressions, and where earlier mentions can provide useful semantic cues. Ignoring these document-level dependencies can lead to inconsistent predictions within the same document~\cite{schwartz_simple_2002}.

\begin{figure}
\centering
\includegraphics[width=\linewidth]{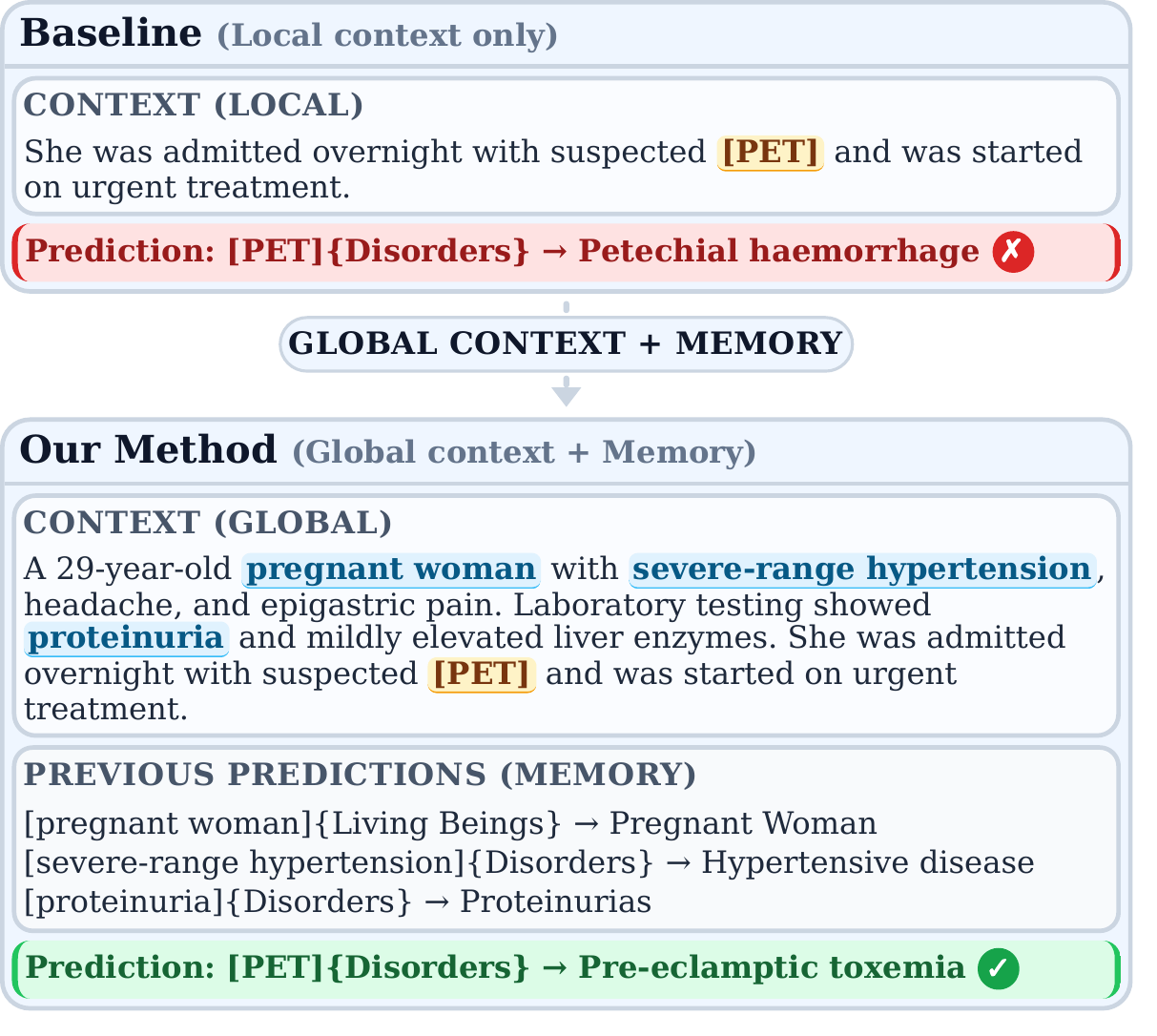}
\caption{Illustrative example of document-level context and memory. \textbf{Top:} Restricted to local context, a standard model mislinks \textit{[PET]} to \textit{Petechial haemorrhage}. \textbf{Bottom:} By using global clinical context and previous predictions, LongBEL correctly links \textit{[PET]} to \textit{Pre-eclamptic toxemia}.}
\label{fig:consistency}
\end{figure}

Collective entity linking has shown that linking decisions can benefit from dependencies between mentions in the same document~\citep{ratinov_local_2011,ganea_deep_2017,agarwal_entity_2022}. However, constrained generative BEL is still typically applied independently to each mention using only local context. As illustrated in Figure~\ref{fig:consistency}, useful document-level evidence may come from repeated concepts or related clinical cues.

We propose \textbf{LongBEL}, a document-level generative framework for BEL. LongBEL combines full-document context with a robust memory of previous predictions that stores previous linking decisions to improve consistency across a document.

Our main contributions are:
\begin{itemize}
    \item We design a robust memory of previous predictions that reuses previous linking decisions while limiting cascading errors.
    \item We combine memory with long-context generation and ensemble decoding, obtaining the best results on five BEL benchmarks across English, French, and Spanish, while comparing against reproduced techniques from recent BEL work.
    \item We provide a detailed analysis showing that document-level information helps most for recurring concepts, and that this effect varies across datasets.
\end{itemize}

All code, preprocessing scripts, and evaluation scripts needed to reproduce the experiments are available in a \href{https://github.com/Aremaki/LongBEL}{GitHub
 repository}. Trained model checkpoints, processed datasets, and the resources required to run inference are available in a \href{https://huggingface.co/collections/Aremaki/longbel}{Hugging
 Face repository}.

\section{Related Work}

We organize prior BEL methods according to the information used for linking: the mention alone, its local context, or dependencies across the document. These categories also correspond to the main baseline families used in our experiments.

\subsection{Context-Free BEL}

Context-free BEL links each mention without using its surrounding context.

\paragraph{Rule-Based and Lexical Systems.}
Early systems such as MetaMap~\citep{aronson_overview_2010}, cTAKES~\citep{savova_mayo_2010}, and the scispaCy UMLS linker~\citep{neumann_scispaCy_2019} rely mainly on dictionary and lexical matching. They are efficient, but struggle with unlisted lexical variants.

\paragraph{Self-Supervised Bi-Encoders.}

Models such as SapBERT~\citep{liu_self-alignment_2021}, CODER~\citep{yuan_coder_2022}, and BERGAMOT~\citep{sakhovskiy_biomedical_2024} learn concept representations from KB synonyms using contrastive learning. They capture lexical variation better than string matching, but cannot use context to resolve ambiguous mentions.

\subsection{Supervised Local-Context BEL}

Local-context models use the sentence or neighboring text around the target mention.

\paragraph{Contextualized Bi-Encoders.}
BLINK~\citep{wu_scalable_2020} uses a bi-encoder to retrieve candidates from representations of the mention context and entity descriptions, followed by cross-encoder reranking.

\paragraph{Generative Models.}
GENRE formulates entity linking as constrained autoregressive generation~\citep{cao_autoregressive_2021}. Biomedical systems such as GenBioEL and ANGEL adapt this formulation to biomedical KBs~\citep{yuan_generative_2022,kim_learning_2025}. Decoder-only language models can also be fine-tuned as generative linkers~\citep{remaki_syncabel_2026}. However, these supervised generative approaches usually link mentions independently using local context.

\subsection{Document-Level Modeling and Consistency}

\paragraph{Collective Entity Linking.}
Collective entity linking exploits dependencies between mentions in the same document. Early approaches combined local predictions with coherence features between candidate entities~\citep{ratinov_local_2011}. Later neural methods combined local and document-level evidence through a CRF, candidate graphs, recurrent random walks, or joint document encoding~\citep{ganea_deep_2017,cao_neural_2018,xue_neural_2019,bhowmik_fast_2021}. In BEL, ArboEL explicitly models mention--mention and mention--entity relations through an arborescence objective~\citep{agarwal_entity_2022}. We therefore use it as our main collective BEL baseline.

\paragraph{Long-Context Models and Memory.}
Most collective approaches are encoder-based. Recent open-weight language models support much longer inputs, making supervised document-level generation increasingly practical~\citep{grattafiori_llama_2024}. Memory mechanisms have also been used to track entities across long narratives~\citep{zhu_llmlink_2025}, but have not been directly explored for constrained generative BEL.

\subsection{Other Uses of Large Language Models for BEL}
When large language models are not fine-tuned as linkers, they are commonly used as auxiliary components, including rerankers~\citep{ding_chatel_2024,wang_aelc_2025,shlyk_real_2024,lin_guiding_2025}, prompting-based disambiguators~\citep{borchert_improving_2024,vollmers_contextual_2025}, and data generators~\citep{xin_llmael_2025,remaki_syncabel_2026}.

\subsection{Positioning of LongBEL}
LongBEL combines supervised generative BEL with collective entity linking. Unlike existing generative linkers, which usually resolve each mention independently from local context, it uses the full document and previous predictions. Unlike existing collective methods, which are mainly encoder-based, LongBEL models cross-mention dependencies through constrained autoregressive generation. Mentions are processed sequentially, and earlier linking decisions are added directly to the input to guide later predictions.

\section{Our Methods}

\subsection{Problem Statement}

Let $\mathcal{E}$ denote the set of candidate concepts in a predefined KB, such as UMLS or SNOMED CT. Each concept $e \in \mathcal{E}$ has a semantic group $g_e$ (e.g., \textit{Disorders}) and a set of synonyms $S_e$. Given a document $c$ (the \textit{global context}) containing an ordered sequence of mentions $\mathcal{M} = (m_1, m_2, \dots, m_n)$, where each mention $m_t$ is associated with a semantic group $g_t$, the goal is to map each mention $m_t$ to its target concept $e_t \in \mathcal{E}$. For example, in Figure~\ref{fig:consistency}, the mention \textit{PET} belongs to the \textit{Disorders} semantic group and should be linked to the concept C0032914, whose synonyms include \textit{Pre-eclamptic toxemia}.

\subsection{Document-level Autoregressive Formulation with Memory}

We frame this entity linking task as a sequence-to-sequence problem, extending the autoregressive formulation of \citet{cao_autoregressive_2021} to a document-level setting. To disambiguate the $t$-th mention $m_t$, our model relies not only on the global context $c$, but also on a \textit{memory} $\mathcal{H}_t$ of concepts predicted earlier in the document.

\paragraph{Memory Mechanism.}
Concepts within the same document are often related or repeated. To use this information, LongBEL keeps a memory of previous linking decisions. At step $t$, the memory is defined as:
\begin{equation*}
\mathcal{H}_t = \{(m_1, e_1), \dots, (m_{t-1}, e_{t-1})\}.
\end{equation*}
For example, before linking the mention \textit{PET} in Figure~\ref{fig:consistency}, the memory may already contain earlier predictions such as (\textit{pregnant woman}, \textit{Pregnant Woman}), (\textit{severe-range hypertension}, \textit{Hypertensive disease}), (\textit{proteinuria}, \textit{Proteinurias}).

\paragraph{Input Format.}
For the $t$-th mention, the input sequence $x_t$ is constructed by concatenating three components:
(i) the full document $c$, 
(ii) the memory $\mathcal{H}_t$ (formatted as a list of previous concepts), and 
(iii) the current target mention $m_t$ alongside its semantic group $g_t$.
In the same example, the target mention is \textit{PET}, with semantic group \textit{Disorders}, and the input also includes the full clinical paragraph and the previous memory entries.

% \vspace{0.2cm}
% \noindent\textit{\textbf{Example:}}
% \vspace{0.1cm}

% \noindent\textit{Input $x_t$:}
% \vspace{0.1cm}
% \small
% \noindent ``\#\#\# Context \\
% \noindent The patient presented with a severe [headache] and was advised to take [aspirin] for the [pain]. \\
% \#\#\# Previous Normalizations \\
% \noindent [headache]\{Disorders\} Headache \\
% \noindent [aspirin]\{Chemicals \& Drugs\} Aspirin \\
% \#\#\# Prediction \\
% \noindent [pain]\{Disorders\}''
% \normalsize
% \vspace{0.2cm}

% \noindent\textit{Target $y_t$:}
% \vspace{0.1cm}
% \small
% \noindent ``Headache'' 
% \normalsize
% \vspace{0.2cm}

\subsection{Training}

\paragraph{Training-Target Selection.}
For each training mention $m_t$ with gold concept $e_t$, the target text $y_t$ is selected from the retained synonym set $S_{e_t}$. Following GenBioEL~\citep{yuan_generative_2022}, we rank these synonyms by their TF-IDF similarity to the mention surface form and select the highest-scoring one. This yields a target string close to the observed mention while preserving a deterministic mapping to the gold KB concept.

\paragraph{Training Objective.}
The model is trained with teacher forcing to generate the target concept text $y_t$ conditioned on the input $x_t$:
\begin{equation*}
p_\theta(y_t \mid x_t) = \prod_i p_\theta(y_{t,i} \mid y_{t,<i}, x_t),
\end{equation*}
where $y_{t,i}$ is the $i$-th token of the target string. This objective allows the model to use the full document context and memory $\mathcal{H}_t$ when predicting the current concept.

\begin{figure*}
    \centering
    \includegraphics[width=0.95\textwidth]{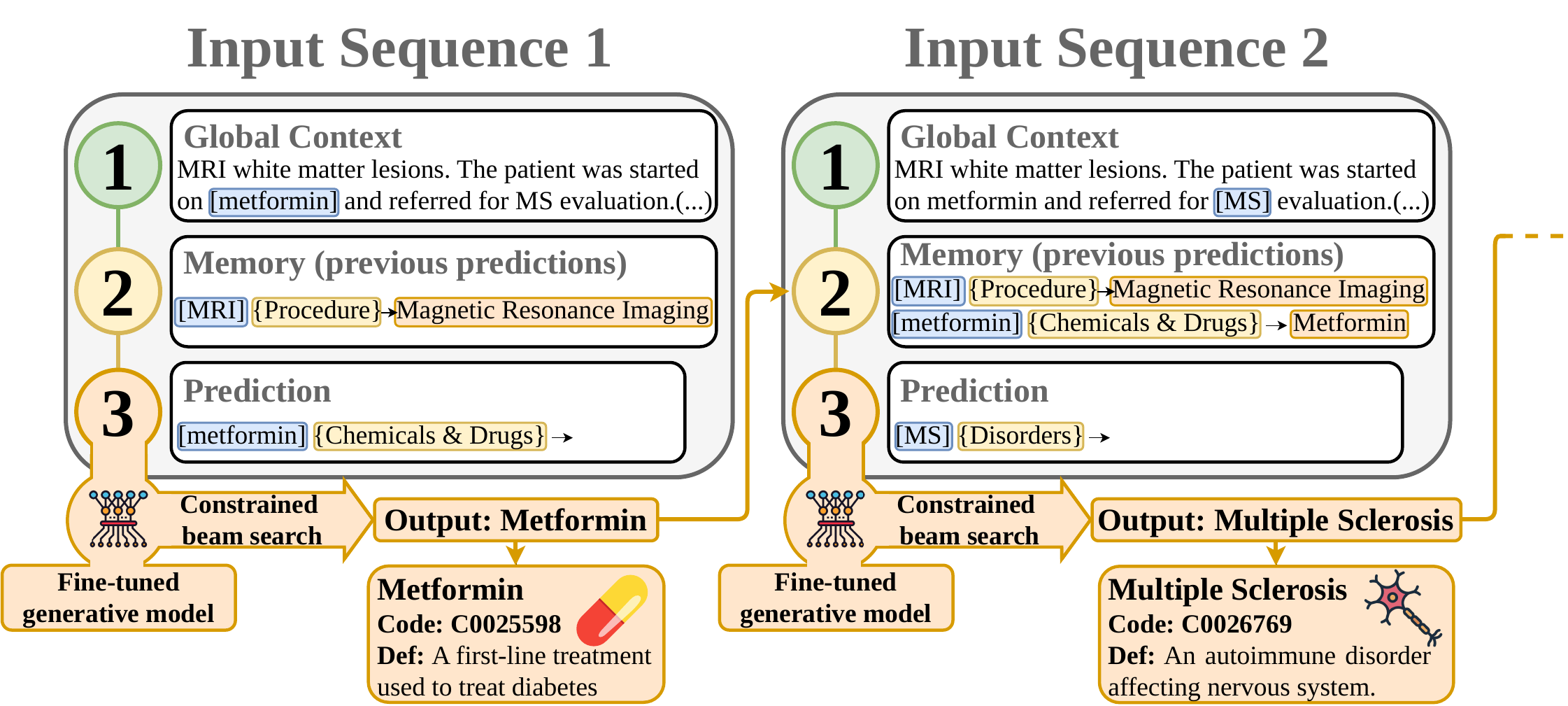}
    \caption{Overview of LongBEL sequential inference. For each mention, the input combines the full document context, a memory of previous predictions, and the current mention with its semantic group. Constrained beam search over the KB-derived trie generates a concept name, which is mapped to its KB identifier and appended to the memory before processing the next mention.}
    \label{fig:inference}
\end{figure*}

\paragraph{Robust Memory Construction via Cross-Validation.}
\label{sec:memory}
During inference, $\mathcal{H}_t$ is filled with the model's own previous predictions, which may be wrong. If the model is trained only with gold concepts in memory, it can over-trust this signal and propagate early errors to later mentions. To reduce this exposure bias, we build the training memory using model predictions instead of gold labels. We use 5-fold cross-validation on the training set: five base models are trained, each on four folds, and each model generates predictions for its held-out fold. These out-of-fold predictions are then used to populate $\mathcal{H}_t$ in the final training data. This exposes LongBEL to realistic inference-time noise and encourages it to balance memory with the current document context. Figure~\ref{fig:robust_memory} in Appendix~\ref{sec:appendix_robust_memory} illustrates this process.

\subsection{Inference}

Inference combines sequential prediction with dynamic memory, constrained decoding over a predefined KB, and an optional ensemble of complementary input configurations.

\paragraph{Sequential Prediction with Dynamic Memory.} As illustrated in Figure~\ref{fig:inference}, LongBEL resolves mentions sequentially in their order of appearance in the document $c$. For the first mention $m_1$, the memory $\mathcal{H}_1$ is empty. After predicting concept $e_t$ for mention $m_t$, the pair $(m_t,e_t)$ is appended to the memory, yielding $\mathcal{H}_{t+1}$. Previous linking decisions can therefore inform later predictions, promoting document-level consistency.

\paragraph{KB-Constrained Decoding.} We assume a predefined candidate KB containing all concepts that the model may predict. Because LongBEL generates concept names rather than KB identifiers, we construct the decoding inventory from the synonyms provided by the KB. We remove only synonym strings associated with multiple concept identifiers within the same semantic group, so that each retained string maps unambiguously to one KB concept; no concepts are removed, as each remains represented by at least one retained name. The resulting inventory is converted into a prefix trie. Following GENRE~\cite{cao_autoregressive_2021} and its semantic-group adaptation in SynCABEL~\cite{remaki_syncabel_2026}, decoding for mention $m_t$ is restricted to concepts belonging to its semantic group $g_t$. At each generation step, only tokens corresponding to valid trie continuations are allowed, ensuring that each completed output maps to a valid and unique KB concept.

\paragraph{Ensemble via Reciprocal Rank Fusion.}
In addition to the individual LongBEL configurations, we evaluate an ensemble combining local-context, global-context, memory-only, and global-context-with-memory variants to test whether they provide complementary signals. Each variant produces a ranked list of candidate concepts using constrained beam search. These lists are combined with Reciprocal Rank Fusion (RRF)~\cite{cormack_reciprocal_2009}. For each candidate concept $e$, the RRF score is:
\begin{equation*}
\operatorname{RRF}(e)=\sum_{v\in\mathcal{V}}\frac{1}{r_v(e)},
\end{equation*}
where $\mathcal{V}$ is the set of variants and $r_v(e)$ is the rank of $e$ under variant $v$. Ranks start at 1, and candidates absent from a ranked list receive a contribution of zero. The candidate with the highest RRF score is selected as the final prediction. Figure~\ref{fig:ensemble_inference} in Appendix~\ref{app:ensemble_inference} illustrates the complete ensemble procedure. As the ensemble requires predictions from multiple configurations for each mention, it incurs a higher inference cost than individual variants.

\section{Experiments}

\subsection{Datasets}

We evaluate LongBEL on five datasets drawn from three annotated corpora spanning three languages: \textbf{MedMentions-ST21pv (MM-ST21pv)}~\cite{mohan_medmentions_2019}, which consists of English PubMed abstracts; \textbf{QUAERO-EMEA}~\cite{neveol_quaero_2014}, comprising French regulatory documents; and the \textbf{Spanish Clinical Case Corpus (SPACCC)}. SPACCC is a single shared collection of 1,000 real-world clinical reports that provides three independent annotations, which we evaluate as separate datasets: \textbf{SympTEMIST} for symptoms~\cite{miranda-escalada_overview_2022}, \textbf{DisTEMIST} for diseases~\cite{lima-lopez_overview_2023}, and \textbf{MedProcNER} for procedures~\cite{lima-lopez_overview_2023-1}.

These corpora were selected because they represent large human-annotated BEL benchmarks in their respective languages. We evaluate the EMEA subset of QUAERO and omit QUAERO-MEDLINE, as the latter contains only short title-length texts and is not suitable for document-level context evaluation. Detailed dataset statistics are reported in Table~\ref{tab:dataset-summary} in Appendix~\ref{sec:dataset_statistics}. The datasets differ strongly in concept recurrence: MM-ST21pv and QUAERO-EMEA have high recurrence rates ($51.5\%$ and $58.8\%$ of mentions correspond to recurring concepts), while the SPACCC clinical datasets have much lower rates ($11.2\%$, $15.4\%$, and $16.3\%$). This variation allows us to evaluate LongBEL under different levels of concept recurrence.

% \begin{table}[h]
% \centering
% \small
% \setlength{\tabcolsep}{1.5pt}
% \renewcommand{\arraystretch}{0.9}
% \begin{tabular}{@{}p{1.7cm}
%                 >{\centering\arraybackslash}p{1.0cm}
%                 >{\centering\arraybackslash}p{1.2cm}
%                 >{\centering\arraybackslash}p{1.1cm}
%                 >{\centering\arraybackslash}p{1.1cm}
%                 >{\centering\arraybackslash}p{1.1cm}@{}}
% \toprule
%  & MM-ST21pv
%  & QUAERO-EMEA
%  & Symp-TEMIST
%  & Dis-TEMIST
%  & MedProc-NER \\
% \midrule
% \# Docs        & 4,392   & 26     & 1,000  & 1,000  & 1,000  \\
% Avg. Len.      & 1,581   & 5,062  & 2,337  & 2,337  & 2,337  \\
% Language       & En.     & Fr.    & Es.    & Es.    & Es.    \\
% Clinical       & \xmark  & \xmark & \cmark & \cmark & \cmark \\
% \# Mentions    & 203,185 & 7,127  & 11,683 & 7,041  & 8,379  \\
% \# Concepts    & 25,419  & 1,226  & 4,161  & 3,015  & 2,351  \\
% \# Sem. Grps   & 14      & 10     & 1      & 1      & 1      \\
% \% First       & 48.5\%  & 41.2\% & 88.8\% & 84.6\% & 83.7\% \\
% \% Recurring    & 51.5\%  & 58.8\% & 11.2\% & 15.4\% & 16.3\% \\
% \bottomrule
% \end{tabular}
% \caption{
% Summary statistics of the evaluation datasets. 
% \# Sem. Grps: number of semantic groups; 
% Avg. Len.: average document length in characters; 
% Clinical: whether the dataset consists of clinical reports; 
% \% First: percentage of mentions whose concept appears for the first time in a document; 
% \% Recurring: percentage of mentions whose concept has already appeared at least once in the same document.
% }
% \label{tab:dataset-summary}
% \end{table}

\subsection{Knowledge Bases}
Each dataset is annotated against a predefined KB specified by its original annotation scheme. We use UMLS 2017AA restricted to the ST21-pv subset for MM-ST21pv, UMLS 2014AA restricted to the 10 relevant semantic groups for QUAERO-EMEA, and task-specific gazetteers derived from the 2021 Spanish SNOMED CT release for SPACCC. Detailed statistics for these KB subsets are reported in Table~\ref{tab:kb_stats} in Appendix~\ref{sec:appendix_kb}.

\subsection{Implementation Details}

We implement LongBEL with PyTorch and Hugging Face Transformers, using one NVIDIA H100 80GB GPU per run. Training and inference hyperparameters are reported in Appendix~\ref{sec:appendix_implementation}.

\begin{table*}[ht]
\centering
\begin{tabular}{lc c c c c}
\toprule
\textbf{Model} &
\shortstack{\textbf{MM-}\\\textbf{ST21PV}\\(English)} &
\shortstack{\textbf{QUAERO-}\\\textbf{EMEA}\\(French)} &
\shortstack{\textbf{Symp-}\\\textbf{TEMIST}\\(Spanish)} &
\shortstack{\textbf{Dis-}\\\textbf{TEMIST}\\(Spanish)} &
\shortstack{\textbf{MedProc-}\\\textbf{NER}\\(Spanish)} \\
\midrule
\multicolumn{6}{l}{\textsc{Context-Free BEL}} \\
\midrule
scispaCy\footnotesize~\cite{neumann_scispaCy_2019} & $53.8 {\scriptstyle \pm 1.0}$ & $37.1 {\scriptstyle \pm 4.3}$ & $9.8 {\scriptstyle \pm 1.3}$ & $21.1 {\scriptstyle \pm 1.9}$ & $10.3 {\scriptstyle \pm 1.2}$ \\
SapBERT\footnotesize~\cite{liu_self-alignment_2021} & $65.6 {\scriptstyle \pm 1.0}$ & $59.7 {\scriptstyle \pm 3.8}$ & $34.2 {\scriptstyle \pm 2.0}$ & $38.6 {\scriptstyle \pm 2.6}$ & $30.4 {\scriptstyle \pm 2.1}$ \\
CODER-all\footnotesize~\cite{yuan_coder_2022} & $62.9 {\scriptstyle \pm 1.1}$ & $66.9 {\scriptstyle \pm 4.0}$ & $42.2 {\scriptstyle \pm 2.2}$ & $47.0 {\scriptstyle \pm 2.6}$ & $42.7 {\scriptstyle \pm 2.1}$ \\
SapBERT-all\footnotesize~\cite{liu_learning_2021} & $64.6 {\scriptstyle \pm 1.1}$ & $67.9 {\scriptstyle \pm 3.9}$ & $49.8 {\scriptstyle \pm 2.4}$ & $49.6 {\scriptstyle \pm 2.6}$ & $45.1 {\scriptstyle \pm 2.2}$ \\
BERGAMOT\footnotesize~\cite{sakhovskiy_biomedical_2024} & $60.9 {\scriptstyle \pm 1.1}$ & $63.8 {\scriptstyle \pm 4.9}$ & $48.0 {\scriptstyle \pm 2.7}$ & $48.9 {\scriptstyle \pm 2.4}$ & $42.3 {\scriptstyle \pm 2.2}$ \\
\midrule
\multicolumn{6}{l}{\textsc{Collective BEL}} \\
\midrule
ArboEL\footnotesize~\cite{agarwal_entity_2022} & $76.9 {\scriptstyle \pm 0.9}$ & $63.0 {\scriptstyle \pm 3.9}$ & $55.4 {\scriptstyle \pm 2.5}$ & $54.7 {\scriptstyle \pm 2.6}$ & $59.7 {\scriptstyle \pm 2.6}$ \\
\midrule
\multicolumn{6}{l}{\textsc{Local-Context Generative BEL}} \\
\midrule
GENRE\footnotesize~\cite{cao_autoregressive_2021} \\
%\quad \textit{mBART-large}\footnotesize~\cite{tang_multilingual_2020} & $69.6 {\scriptstyle \pm 1.0}$ & $69.3 {\scriptstyle \pm 5.4}$ & $59.8 {\scriptstyle \pm 2.7}$ & $58.7 {\scriptstyle \pm 2.7}$ & $66.0 {\scriptstyle \pm 2.3}$ \\
\quad \textit{Llama-1B}\footnotesize~\cite{grattafiori_llama_2024} & $73.1 {\scriptstyle \pm 1.0}$ & ${75.1} {\scriptstyle \pm 3.6}$ & $60.5 {\scriptstyle \pm 2.4}$ & $62.5 {\scriptstyle \pm 2.3}$ & $67.4 {\scriptstyle \pm 2.1}$ \\
\quad \textit{Llama-8B}\footnotesize~\cite{grattafiori_llama_2024} & $75.0 {\scriptstyle \pm 0.9}$ & $73.8 {\scriptstyle \pm 4.0}$ & ${61.7} {\scriptstyle \pm 2.5}$ & ${63.2} {\scriptstyle \pm 2.5}$ & ${68.3} {\scriptstyle \pm 2.2}$ \\
\midrule
\multicolumn{6}{l}{\textsc{Generative Collective BEL (Our Methods)}} \\
\midrule
LongBEL-1B & $77.6 {\scriptstyle \pm 0.9}$ & $74.5 {\scriptstyle \pm 3.7}$ & $59.8 {\scriptstyle \pm 2.5}$ & $61.9 {\scriptstyle \pm 2.4}$ & $66.6 {\scriptstyle \pm 2.1}$ \\
LongBEL-8B & $\underline{79.3} {\scriptstyle \pm 0.8}$ & $75.4 {\scriptstyle \pm 3.4}$ & $62.0 {\scriptstyle \pm 2.6}$ & $63.6 {\scriptstyle \pm 2.1}$ & $69.0 {\scriptstyle \pm 2.1}$ \\
\midrule
LongBEL-1B-Ensemble & $79.1 {\scriptstyle \pm 0.8}$ & $\underline{77.2} {\scriptstyle \pm 3.0}$ & $\underline{62.3} {\scriptstyle \pm 2.4}$ & $\underline{64.7} {\scriptstyle \pm 2.2}$ & $\underline{69.7} {\scriptstyle \pm 2.1}$ \\
LongBEL-8B-Ensemble & $\textbf{80.6} {\scriptstyle \pm 0.8}$ & $\textbf{77.6} {\scriptstyle \pm 3.0}$ & $\textbf{63.6} {\scriptstyle \pm 2.5}$ & $\textbf{65.8} {\scriptstyle \pm 2.2}$ & $\textbf{71.0} {\scriptstyle \pm 2.0}$ \\
\bottomrule
\end{tabular}
\caption{
Entity linking performance (Recall@1) on biomedical benchmarks. \textbf{LongBEL} combines full-document context with a robust memory of previous predictions. ``Ensemble'' denotes Reciprocal Rank Fusion over four input configurations. Best results are in \textbf{bold}, and second-best results are \underline{underlined}. The 95\% confidence intervals are estimated using empirical bootstrap resampling at the document level.
}
\label{tab:benchmark-results}
\end{table*}

\subsection{Results}

\subsubsection{Benchmark Performance}
\label{sec:benchmark_performance}

Table~\ref{tab:benchmark-results} reports Recall@1 scores with $\pm$ half the 95\% confidence interval across five BEL benchmarks. We organize the baselines into three families. Context-free BEL includes the scispaCy UMLS linker and bi-encoder methods that resolve mentions independently. ArboEL is our collective BEL baseline and explicitly models mention--mention relations to encourage consistency. GENRE represents local-context generative BEL and predicts each concept from the sentence containing the target mention. This organization reflects the positioning of LongBEL at the intersection of generative and collective BEL.

To isolate document-level modeling, we compare GENRE and LongBEL with the same Llama-1B and Llama-8B backbones and constrained generative formulation. GENRE uses only the target sentence, whereas LongBEL also uses the full document and a robust memory of previous predictions. The comparison between the corresponding GENRE and LongBEL models therefore measures the effect of adding document context and cross-mention memory while keeping the rest fixed.

LongBEL-8B is the best single model on all five benchmarks. Compared with GENRE-Llama-8B, it gains 4.3 Recall@1 points on MM-ST21pv, 1.6 on QUAERO-EMEA, and 0.3--0.7 across the three SPACCC datasets. These results indicate that the benefit of document-level information varies across datasets, with the clearest improvement observed on MM-ST21pv.

We additionally evaluate an ensemble using RRF over four configurations: local context only, global context only, local context with robust memory, and global context with robust memory. They achieve the best results on every dataset, indicating that these signals are complementary.

To ensure a comparable candidate space, all baselines were reproduced and evaluated under the same setting. They use the same KBs and candidate concepts are restricted to the semantic group of each mention. The only exception is the scispaCy UMLS linker, which operates over the full UMLS inventory and does not support semantic-group filtering.

\begin{table}
\centering
\small
\setlength{\tabcolsep}{1.5pt}
\renewcommand{\arraystretch}{1.0}
\begin{tabular}{@{} 
                >{\centering\arraybackslash}m{0.7cm}
                >{\centering\arraybackslash}m{0.7cm}
                c c c c c @{}}
\toprule
\textbf{Ctx} & \textbf{Mem.} 
 & \shortstack{MM}
 & \shortstack{EMEA}
 & \shortstack{Symp-\\TEMIST}
 & \shortstack{Dis-\\TEMIST}
 & \shortstack{MedProc-\\NER} \\
\midrule
\multicolumn{7}{c}{\textbf{Llama-1B}} \\ 
\midrule
\xmark & \xmark & $73.1 {\scriptstyle \pm 1.0}$ & $75.1 {\scriptstyle \pm 3.6}$ & $\underline{60.5} {\scriptstyle \pm 2.4}$ & $\underline{62.5} {\scriptstyle \pm 2.3}$ & $67.4 {\scriptstyle \pm 2.1}$ \\
\cmark & \xmark & $76.5 {\scriptstyle \pm 0.9}$ & $
\underline{76.0} {\scriptstyle \pm 2.8}$ & $60.3 {\scriptstyle \pm 2.6}$ & $62.0 {\scriptstyle \pm 2.3}$ & $\underline{67.5} {\scriptstyle \pm 2.1}$ \\
\xmark & \cmark & $77.4 {\scriptstyle \pm 0.8}$ & $74.0 {\scriptstyle \pm 3.2}$ & $60.1 {\scriptstyle \pm 2.7}$ & $62.0 {\scriptstyle \pm 2.3}$ & $67.2 {\scriptstyle \pm 2.1}$ \\
\cmark & \cmark & $\underline{77.6} {\scriptstyle \pm 0.9}$ & $74.5 {\scriptstyle \pm 3.7}$ & $59.8 {\scriptstyle \pm 2.5}$ & $61.9 {\scriptstyle \pm 2.4}$ & $66.6 {\scriptstyle \pm 2.1}$ \\
\addlinespace 
\multicolumn{2}{@{}l}{\textbf{Ensemble}} & $\textbf{79.1} {\scriptstyle \pm 0.8}$ & $\textbf{77.2} {\scriptstyle \pm 3.0}$ & $\textbf{62.3} {\scriptstyle \pm 2.4}$ & $\textbf{64.7} {\scriptstyle \pm 2.2}$ & $\textbf{69.7} {\scriptstyle \pm 2.1}$ \\
\midrule
\multicolumn{7}{c}{\textbf{Llama-8B}} \\ 
\midrule
\xmark & \xmark & $75.0 {\scriptstyle \pm 0.9}$ & $73.8 {\scriptstyle \pm 4.0}$ & $61.7 {\scriptstyle \pm 2.5}$ & $63.2 {\scriptstyle \pm 2.5}$ & $68.3 {\scriptstyle \pm 2.2}$ \\
\cmark & \xmark & $78.6 {\scriptstyle \pm 0.9}$ & $73.9 {\scriptstyle \pm 3.1}$ & $61.3 {\scriptstyle \pm 2.6}$ & $63.3 {\scriptstyle \pm 2.3}$ & $\underline{69.1} {\scriptstyle \pm 2.2}$ \\
\xmark & \cmark & $79.0 {\scriptstyle \pm 0.9}$ & $75.3 {\scriptstyle \pm 3.0}$ & $61.8 {\scriptstyle \pm 2.4}$ & $62.5 {\scriptstyle \pm 2.4}$ & $68.2 {\scriptstyle \pm 2.3}$ \\
\cmark & \cmark & $\underline{79.3} {\scriptstyle \pm 0.8}$ & $\underline{75.4} {\scriptstyle \pm 3.4}$ & $ \underline{62.0} {\scriptstyle \pm 2.6}$ & $\underline{63.6} {\scriptstyle \pm 2.1}$ & $69.0 {\scriptstyle \pm 2.1}$ \\
\addlinespace
\multicolumn{2}{@{}l}{\textbf{Ensemble}} & $\textbf{80.6} {\scriptstyle \pm 0.8}$ & $\textbf{77.6} {\scriptstyle \pm 3.0}$ & $\textbf{63.6} {\scriptstyle \pm 2.5}$ & $\textbf{65.8} {\scriptstyle \pm 2.2}$ & $\textbf{71.0} {\scriptstyle \pm 2.0}$ \\
\bottomrule
\end{tabular}
\caption{Ablation study (recall) evaluating LongBEL-1B and LongBEL-8B. `Ctx' denotes the inclusion of global context, and `Mem.' indicates whether memory is enabled (\cmark) or disabled (\xmark). Best results are in \textbf{bold}, second-best are \underline{underlined}. 95\% confidence intervals are estimated using empirical bootstrap resampling at the document level. The ensemble rows fuse all components via RRF.}
\label{tab:ablation}
\end{table} 

\subsubsection{Ablation Study}

To better understand the role of global context (Ctx) and robust memory (Mem), we conduct an ablation study with both LongBEL-1B and LongBEL-8B, as reported in Table~\ref{tab:ablation}. On MM-ST21pv, both global context and memory help improve performance over the baseline. The best single-model results are obtained when the two components are used together, suggesting that both sources of document-level information are useful on this dataset. Global context allows the model to use evidence from the whole document, while memory gives it access to previous linking decisions that may help maintain consistency across mentions.

On the remaining datasets, improvements are more moderate and less consistent, with differences often within confidence intervals. This was already observed in the main benchmark results and suggests that the usefulness of document-level information varies across benchmarks. We also observe that larger models (8B) integrate global context and memory more reliably, whereas smaller models (1B) exhibit less stable behavior when combining these components.

Finally, in the main Recall@1 comparison, the ensemble obtains the best result on each dataset. This suggests that local context, global context, and memory provide complementary signals, leading to more robust performance when combined. However, this requires four model runs and therefore substantially increases inference cost.

\section{Discussion}

\subsection{Impact of Concept Redundancy}

LongBEL’s gains come mainly from enforcing consistency across concepts that appear more than once in a document. Table~\ref{tab:consistency-table} separates first occurrences from recurring mentions. For the 8B models, LongBEL differs from GENRE by only $-0.4$ to $+0.6$ points on first occurrences, but gains $1.4$--$8.3$ points on recurring mentions. The largest gain is observed on MM-ST21pv ($+8.3$). The pattern is less consistent for the 1B models, although again the largest gain is on recurring mentions in MM-ST21pv ($+8.6$).

Recurrence helps explain the larger gains on MM-ST21pv than on SPACCC. MM-ST21pv contains $51.5\%$ recurring mentions, whereas the SPACCC datasets contain only $11.2$--$16.3\%$, providing fewer opportunities to reuse earlier predictions. However, QUAERO-EMEA is an exception: despite its high recurrence, gains remain modest. Recurrence creates opportunities to use memory at inference, but learning when to trust memory requires sufficient training data. QUAERO-EMEA has only 15 training documents, leaving several semantic groups sparsely represented. The group-level results in Appendix~\ref{sec:appendix_sem_group} are consistent with this hypothesis: gains tend to be larger for groups with more training examples and more variable for sparse ones.

Appendix~\ref{sec:appendix_detailed_subset_results} reports further subset analyses, while Appendix~\ref{sec:appendix_qual_examples} illustrates when memory helps or propagates errors.

\begin{table}[t]
\small
\centering
\setlength{\tabcolsep}{2.5pt}
\renewcommand{\arraystretch}{1.0}
\begin{tabular}{l c c c c}
\toprule
& \multicolumn{2}{c}{\textbf{Llama-1B}} & \multicolumn{2}{c}{\textbf{Llama-8B}} \\
\cmidrule(lr){2-3} \cmidrule(lr){4-5}
\textbf{Subset} & Baseline & LongBEL & Baseline & LongBEL \\
\midrule
\multicolumn{5}{l}{\textsc{MedMentions-ST21pv}} \\
\midrule
Overall  & $73.1 {\scriptstyle \pm 1.0}$ & $77.6 {\scriptstyle \pm 0.9}$ & $75.0 {\scriptstyle \pm 0.9}$ & $79.3 {\scriptstyle \pm 0.8}$ \\
First   & $79.1 {\scriptstyle \pm 0.7}$ & $79.1 {\scriptstyle \pm 0.7}$ & $80.2 {\scriptstyle \pm 0.7}$ & $80.3 {\scriptstyle \pm 0.7}$ \\
Recurring & $67.5 {\scriptstyle \pm 1.4}$ & $76.1 {\scriptstyle \pm 1.4}$ & $70.1 {\scriptstyle \pm 1.5}$ & $78.4 {\scriptstyle \pm 1.3}$ \\
\midrule
\multicolumn{5}{l}{\textsc{QUAERO-EMEA}} \\
\midrule
Overall  & $75.1 {\scriptstyle \pm 3.6}$ & $74.5 {\scriptstyle \pm 3.7}$ & $73.8 {\scriptstyle \pm 4.0}$ & $75.4 {\scriptstyle \pm 3.4}$ \\
First   & $75.0 {\scriptstyle \pm 3.4}$ & $72.6 {\scriptstyle \pm 3.6}$ & $73.6 {\scriptstyle \pm 3.2}$ & $74.2 {\scriptstyle \pm 3.5}$ \\
Recurring & $75.2 {\scriptstyle \pm 5.4}$ & $75.8 {\scriptstyle \pm 5.2}$ & $74.0 {\scriptstyle \pm 5.6}$ & $76.3 {\scriptstyle \pm 4.2}$ \\
\midrule
\multicolumn{5}{l}{\textsc{SympTEMIST}} \\
\midrule
Overall  & $60.5 {\scriptstyle \pm 2.4}$ & $59.8 {\scriptstyle \pm 2.5}$ & $61.7 {\scriptstyle \pm 2.5}$ & $62.0 {\scriptstyle \pm 2.6}$ \\
First   & $58.0 {\scriptstyle \pm 2.5}$ & $57.6 {\scriptstyle \pm 2.5}$ & $59.6 {\scriptstyle \pm 2.7}$ & $59.8 {\scriptstyle \pm 2.5}$ \\
Recurring & $79.0 {\scriptstyle \pm 5.0}$ & $75.6 {\scriptstyle \pm 5.3}$ & $76.5 {\scriptstyle \pm 5.3}$ & $77.9 {\scriptstyle \pm 5.1}$ \\
\midrule
\multicolumn{5}{l}{\textsc{DisTEMIST}} \\
\midrule
Overall  & $62.5 {\scriptstyle \pm 2.3}$ & $61.9 {\scriptstyle \pm 2.4}$ & $63.2 {\scriptstyle \pm 2.5}$ & $63.6 {\scriptstyle \pm 2.1}$ \\
First   & $61.6 {\scriptstyle \pm 2.5}$ & $60.7 {\scriptstyle \pm 2.4}$ & $62.2 {\scriptstyle \pm 2.7}$ & $61.8 {\scriptstyle \pm 2.2}$ \\
Recurring & $67.4 {\scriptstyle \pm 5.3}$ & $68.6 {\scriptstyle \pm 5.5}$ & $68.9 {\scriptstyle \pm 5.3}$ & $74.3 {\scriptstyle \pm 4.8}$ \\
\midrule
\multicolumn{5}{l}{\textsc{MedProcNER}} \\
\midrule
Overall  & $67.4 {\scriptstyle \pm 2.1}$ & $66.6 {\scriptstyle \pm 2.1}$ & $68.3 {\scriptstyle \pm 2.2}$ & $69.0 {\scriptstyle \pm 2.1}$ \\
First   & $66.2 {\scriptstyle \pm 2.1}$ & $65.1 {\scriptstyle \pm 2.2}$ & $67.1 {\scriptstyle \pm 2.1}$ & $67.4 {\scriptstyle \pm 2.3}$ \\
Recurring & $73.7 {\scriptstyle \pm 4.5}$ & $74.8 {\scriptstyle \pm 4.2}$ & $75.3 {\scriptstyle \pm 4.3}$ & $77.6 {\scriptstyle \pm 3.8}$ \\
\bottomrule
\end{tabular}
\caption{Recall@1 ($\pm$ half the 95\% confidence interval) for mentions whose concept appears for the first time in a document (First) or has appeared previously (Recurring). GENRE uses sentence-level context, whereas LongBEL adds full-document context and robust memory. Intervals are estimated by document-level bootstrap.}
\label{tab:consistency-table}
\end{table}

\subsection{Impact of Robust Memory}

LongBEL uses previous predictions, which can propagate errors. Training with gold memory may make the model over-rely on this signal and copy early mistakes. We therefore train with out-of-fold predictions, as described in Section~\ref{sec:memory}. This substantially increases training cost, requiring five additional training runs and inference over the full training set before the final model training.

Under the same inference conditions, robust memory has limited impact on LongBEL-1B but provides more consistent gains for LongBEL-8B. Recall@1 improves for LongBEL-8B from $78.9$ to $79.3$ on MM-ST21pv, $75.2$ to $75.4$ on QUAERO-EMEA, and $64.5$ to $65.3$ on SPACCC, suggesting that the additional cost is most beneficial for larger models. Detailed results are provided in Table~\ref{tab:memory_training} in Appendix~\ref{sec:appendix_copy_error}.

We also report the \textit{Copy Wrong Memory Error Rate} (CWME), which measures repeated propagation of incorrect predictions. Robust memory reduces CWME for both model sizes, by up to 20.7, 4.5, and 10.5 percentage points on MM-ST21pv, QUAERO-EMEA, and SPACCC, respectively (Figures~\ref{fig:copy_error_1b} and~\ref{fig:copy_error_8b} in Appendix~\ref{sec:appendix_copy_error}). It therefore reduces cascading errors while improving the global precision.

The saliency maps in Figure~\ref{fig:saliency_maps} illustrate these effects: memory helps resolve ambiguous abbreviations (ND and PET), while an incorrect IFN-$\gamma$ normalization propagates to later mentions.

\subsection{Cross-Dataset Generalization}
\label{sec:cross_dataset_generalization}

To evaluate transfer to unseen datasets, we train each model on one dataset and evaluate it directly on another, without target-dataset training examples. At inference, decoding uses the KB associated with the evaluation dataset. This assumes that the target KB is known, but not its annotations. All models use the same Llama-3-8B backbone.

\begin{table}[t]
\centering
\small
\setlength{\tabcolsep}{4pt}
\renewcommand{\arraystretch}{1.08}
\resizebox{\columnwidth}{!}{%
\begin{tabular}{llccc}
\toprule
\shortstack{\textbf{Training}\\\textbf{dataset}} & \textbf{Model}
& \multicolumn{3}{c}{\shortstack{\textbf{Evaluation}\\\textbf{dataset}}} \\
\cmidrule(lr){3-5}
& &
\shortstack{\textbf{MM-}\\\textbf{ST21pv}} &
\shortstack{\textbf{QUAERO-}\\\textbf{EMEA}} &
\textbf{SPACCC} \\
\midrule
\multirow{2}{*}{\shortstack{\textbf{MM-}\\\textbf{ST21pv}}}
& Baseline
& $75.0 {\scriptstyle \pm 0.9}$
& $62.3 {\scriptstyle \pm 4.6}$
& $16.2 {\scriptstyle \pm 1.0}$ \\
& LongBEL
& $\mathbf{79.3} {\scriptstyle \pm 0.8}$
& $\mathbf{72.4} {\scriptstyle \pm 2.7}$
& $\mathbf{25.6} {\scriptstyle \pm 1.4}$ \\
\midrule
\multirow{2}{*}{\shortstack{\textbf{QUAERO-}\\\textbf{EMEA}}}
& Baseline
& $56.4 {\scriptstyle \pm 1.0}$
& $73.8 {\scriptstyle \pm 4.0}$
& $11.8 {\scriptstyle \pm 0.9}$ \\
& LongBEL
& $\mathbf{61.8} {\scriptstyle \pm 1.1}$
& $\mathbf{75.4} {\scriptstyle \pm 3.4}$
& $\mathbf{27.4} {\scriptstyle \pm 1.6}$ \\
\midrule
\multirow{2}{*}{\textbf{SPACCC}}
& Baseline
& $43.6 {\scriptstyle \pm 1.0}$
& $41.4 {\scriptstyle \pm 6.0}$
& $65.1 {\scriptstyle \pm 1.5}$ \\
& LongBEL
& $\mathbf{56.8} {\scriptstyle \pm 1.1}$
& $\mathbf{56.4} {\scriptstyle \pm 3.3}$
& $\mathbf{65.3} {\scriptstyle \pm 1.5}$ \\
\bottomrule
\end{tabular}%
}
\caption{Cross-dataset Recall@1 with target-KB decoding. Models are trained on the row dataset and evaluated on the column dataset without further training. Values are point estimates with half of the 95\% confidence interval.}
\label{tab:cross_dataset_target_kb}
\end{table}

LongBEL has a higher recall@1 than the sentence-level Baseline in all six transfer directions, with gains from $5.4$ to $15.6$ points, showing that document-level context and memory remain useful on unseen datasets. Results for the stricter setting in which the source KB is also transferred unchanged are provided in Table~\ref{tab:cross_dataset_source_kb} in Appendix~\ref{sec:cross_dataset_kb}.

\subsection{Efficiency Trade-offs}

BEL is often deployed at scale, making inference cost an important consideration~\cite{remaki_improving_2025,french_overview_2023}. Table~\ref{tab:efficiency} reports inference speed and memory usage. Among bi-encoders, we include only SapBERT-all since their costs are similar and it provides the best performance.

Context-free methods are fastest but require larger candidate structures. ArboEL's cross-encoder reranking reduces throughput, whereas generative models use compact candidate tries independent of the backbone. LongBEL introduces no additional parameters, so its model memory matches GENRE, but full-document inputs and sequential memory updates reduce speed. To improve throughput, we batch documents during inference, with each document maintaining an independent memory state (Appendix~\ref{sec:inference_parallelization}). Each ensemble combines four models. The reported memory assumes sequential loading, while loading all models simultaneously requires about four times more memory.

\begin{table}
\centering
\resizebox{\columnwidth}{!}{
\begin{tabular}{lccc}
\toprule
 & \shortstack{Model\\(GB)} 
 & \shortstack{Cand.\\(GB)} 
 & \shortstack{Speed\\(ment./s)} \\
\midrule
ScispaCy UMLS Linker & 1.8 & 9.8 & 147.0 \\
SapBERT-all & 2.1 & 20.1 & \textbf{575.5} \\
\midrule
ArboEL & \textbf{1.2} & 7.1 & 38.9 \\
\midrule
%GENRE (mBART-large) & 2.3 & \textbf{5.4} & 51.0 \\
GENRE (Llama-1B) & 2.4 & \textbf{5.4} & 69.6 \\
GENRE (Llama-8B) & 28.6 & \textbf{5.4} & 38.2 \\
\midrule
LongBEL-1B & 2.4 & \textbf{5.4} & 48.5 \\
LongBEL-8B & 28.6 & \textbf{5.4} & 15.2 \\
\midrule
LongBEL-1B-Ensemble & 2.4 & \textbf{5.4} & 14.8 \\
LongBEL-8B-Ensemble & 28.6 & \textbf{5.4} & 6.7 \\
\bottomrule
\end{tabular}
}
\caption{Inference speed and memory footprint. Model: model size; Cand.: candidate memory size; ment.: mentions.}
\label{tab:efficiency}
\end{table}

\subsection{Impact of Semantic Groups}
\label{sec:semantic_groups}

LongBEL assumes that named entity recognition assigns a semantic group to each mention, allowing decoding to be restricted to the corresponding KB subset. Because such labels may not always be available, Appendix~\ref{sec:semantic_group_ablation} compares the Baseline and LongBEL with and without semantic groups. Semantic groups improve absolute Recall@1, particularly on MM-ST21pv ($4.0$--$4.8$ points), but removing them leaves LongBEL's gain over the Baseline broadly unchanged.

\section{Conclusion and Future Work}

We presented \textbf{LongBEL}, a document-level generative framework for BEL that combines full-document context with memory of previous predictions to improve consistency across recurring concepts. Across five biomedical benchmarks, LongBEL improves over sentence-level generative baselines, with the largest gains when concept recurrence is frequent.

Future work will explore two main directions. First, the ensemble results show that local context, global context, and memory provide complementary signals, but this comes with higher inference cost. A practical direction is to apply these components selectively, for example only for ambiguous or recurring mentions. Second, LongBEL is complementary to recent improvements in generative biomedical entity linking. GenBioEL~\cite{yuan_generative_2022} and SynCABEL~\cite{remaki_improving_2025} improve the underlying model through additional pretraining, including the use of generated BEL data, while ANGEL~\cite{kim_learning_2025} introduces contrastive training with negative sampling to improve entity discrimination. Combining these approaches with LongBEL's document-level memory would require methodological and engineering work, but represents a promising direction for future research.

\section{Limitations}

LongBEL has several limitations. First, our experiments assume gold mention spans and semantic groups provided by the NER stage. In a fully end-to-end setting, where a NER model first detects mentions and predicts their semantic groups before BEL is applied, errors from this upstream step could propagate to entity linking. The analysis in Section~\ref{sec:semantic_groups} shows the benefit of semantic groups when they are correct, but does not measure the impact of predicting them. Future work should compare group-constrained BEL using NER-predicted semantic groups with full-KB BEL using span-only NER, to assess whether the reduction in candidate ambiguity compensates for semantic-group prediction errors.

Second, the benefit of LongBEL depends on the structure of the dataset. Our analysis shows that the model is most useful when concepts recur within documents, because memory can then provide useful consistency signals. When most concepts appear only once, memory has little information to exploit, and gains over sentence-level models are smaller. However, recurrence alone is not sufficient. The results on QUAERO-EMEA suggest that even when a high proportion of mentions correspond to recurring concepts, gains may remain moderate if the absolute number of examples per semantic group is small. This means that LongBEL is not expected to provide the same improvement on all biomedical corpora, and is most effective when both recurrence and sample size are sufficient.

Third, LongBEL relies on sequential memory. During inference, each prediction is added to memory and can influence later predictions in the same document. Robust memory training reduces the risk of blindly copying wrong predictions, but it does not fully remove error propagation. If an early prediction is incorrect and later mentions are ambiguous, the model may still reuse the wrong information.

Finally, LongBEL has a higher training and inference cost than sentence-level baselines. It processes longer document contexts, updates memory after each prediction, and therefore cannot parallelize mentions within the same document. However, inference can still be parallelized across documents through batching, with each document maintaining an independent memory state. The ensemble further increases this cost, requiring multiple forward passes per mention and approximately four times more memory than a single LongBEL model. LongBEL is therefore best suited for applications where document-level consistency and linking accuracy are prioritized over maximum throughput.

\section{Ethical Considerations}

This work uses publicly available biomedical entity linking benchmarks. We do not collect new patient data or access private electronic health records. Clinical datasets used in this study are released as de-identified research benchmarks by their original providers.

LongBEL may still produce incorrect normalizations, which could affect downstream clinical applications such as coding, cohort selection, or decision support. It should therefore not be used for clinical decisions without validation, monitoring, and appropriate human oversight.

Our method requires more computation than sentence-level baselines due to document-level inference and robust memory construction. This additional computational cost may increase energy consumption and environmental impact. Future work should explore more efficient training and inference strategies that preserve the benefits of document-level consistency while reducing resource requirements.

All datasets and resources were used according to their licenses and access conditions, including MedMentions, QUAERO, SPACCC, UMLS, and SNOMED CT. The authors used GitHub Copilot for code completion; all generated code was reviewed by the authors.

\section*{Acknowledgments}

This work was performed using HPC resources from GENCI-IDRIS (Grant 2026-AD010316788R1). This work was funded by the Région Île-de-France through the AI4IDF doctoral scholarship and by government funding managed by the French National Research Agency (ANR) under the France 2030 program, grant ANR-23-IACL-0007.

\bibliography{references}

@article{french_overview_2023,
	title = {An overview of {Biomedical} {Entity} {Linking} throughout the years},
	volume = {137},
	issn = {1532-0464},
	url = {https://pmc.ncbi.nlm.nih.gov/articles/PMC9845184/},
	doi = {10.1016/j.jbi.2022.104252},
	urldate = {2026-01-15},
	journal = {Journal of biomedical informatics},
	author = {French, Evan and McInnes, Bridget T.},
	month = jan,
	year = {2023},
	pages = {104252},
}

@inproceedings{bhowmik_fast_2021,
	address = {online},
	title = {Fast and {Effective} {Biomedical} {Entity} {Linking} {Using} a {Dual} {Encoder}},
	url = {https://aclanthology.org/2021.louhi-1.4/},
	urldate = {2026-05-20},
	booktitle = {Proceedings of the 12th {International} {Workshop} on {Health} {Text} {Mining} and {Information} {Analysis}},
	publisher = {Association for Computational Linguistics},
	author = {Bhowmik, Rajarshi and Stratos, Karl and de Melo, Gerard},
	editor = {Holderness, Eben and Jimeno Yepes, Antonio and Lavelli, Alberto and Minard, Anne-Lyse and Pustejovsky, James and Rinaldi, Fabio},
	month = apr,
	year = {2021},
	pages = {28--37},
}

@inproceedings{cao_neural_2018,
  title     = {Neural Collective Entity Linking},
  author    = {Cao, Yixin and Hou, Lei and Li, Juanzi and Liu, Zhiyuan},
  booktitle = {Proceedings of the 27th International Conference on Computational Linguistics},
  year      = {2018},
  pages     = {675--686},
  publisher = {Association for Computational Linguistics},
  url       = {https://aclanthology.org/C18-1057/}
}

@inproceedings{xue_neural_2019,
  title     = {Neural Collective Entity Linking Based on Recurrent Random Walk Network Learning},
  author    = {Xue, Mengge and Cai, Weiming and Su, Jinsong and Song, Linfeng and Ge, Yubin and Liu, Yubao and Wang, Bin},
  booktitle = {Proceedings of the Twenty-Eighth International Joint Conference on Artificial Intelligence},
  year      = {2019},
  pages     = {5327--5333},
  doi       = {10.24963/ijcai.2019/740}
}

@article{remaki_improving_2025,
	title = {Improving {Phenotyping} of {Patients} {With} {Immune}-{Mediated} {Inflammatory} {Diseases} {Through} {Automated} {Processing} of {Discharge} {Summaries}: {Multicenter} {Cohort} {Study}},
	volume = {13},
	copyright = {This is an open-access article distributed under the terms of the Creative Commons Attribution License (https://creativecommons.org/licenses/by/4.0/), which permits unrestricted use, distribution, and reproduction in any medium, provided the original work, first published JMIR Medical Informatics, is properly cited. The complete bibliographic information, a link to the original publication on https://medinform.jmir.org/, as well as this copyright and license information must be included.},
	shorttitle = {Improving {Phenotyping} of {Patients} {With} {Immune}-{Mediated} {Inflammatory} {Diseases} {Through} {Automated} {Processing} of {Discharge} {Summaries}},
	url = {https://medinform.jmir.org/2025/1/e68704},
	doi = {10.2196/68704},
	language = {EN},
	number = {1},
	urldate = {2025-07-11},
	journal = {JMIR Medical Informatics},
	author = {Remaki, Adam and Ung, Jacques and Pages, Pierre and Wajsburt, Perceval and Liu, Elise and Faure, Guillaume and Petit-Jean, Thomas and Tannier, Xavier and Gérardin, Christel},
	month = apr,
	year = {2025},
	note = {Company: JMIR Medical Informatics
Distributor: JMIR Medical Informatics
Institution: JMIR Medical Informatics
Label: JMIR Medical Informatics
Publisher: JMIR Publications Inc., Toronto, Canada},
	pages = {e68704},
}

@inproceedings{neveol_quaero_2014,
	title = {The {QUAERO} {French} {Medical} {Corpus}: {A} {Ressource} for {Medical} {Entity} {Recognition} and {Normalization}},
	booktitle = {Proc of {BioTextMining} {Work}},
	author = {Névéol, Aurélie and Grouin, Cyril and Leixa, Jeremy and Rosset, Sophie and Zweigenbaum, Pierre},
	year = {2014},
	pages = {24--30},
}

@inproceedings{miranda-escalada_overview_2022,
	title = {Overview of {DisTEMIST} at {BioASQ}: {Automatic} detection and normalization of diseases from clinical texts: results, methods, evaluation and multilingual resources},
	url = {https://ceur-ws.org/Vol-3180/paper-11.pdf},
	booktitle = {Working {Notes} of {Conference} and {Labs} of the {Evaluation} ({CLEF}) {Forum}. {CEUR} {Workshop} {Proceedings}},
	author = {Miranda-Escalada, Antonio and Gasco, Luis and Lima-López, Salvador and Farré-Maduell, Eulàlia and Estrada, Darryl and Nentidis, Anastasios and Krithara, Anastasia and Katsimpras, Georgios and Paliouras, Georgios and Krallinger, Martin},
	year = {2022},
}

@inproceedings{lima-lopez_overview_2023,
	title = {Overview of {SympTEMIST} at {BioCreative} {VIII}: corpus, guidelines and evaluation of systems for the detection and normalization of symptoms, signs and findings from text},
	doi = {10.5281/zenodo.10104547},
	author = {Lima-López, Salvador and Farré-Maduell, Eulalia and Gasco Sanchez, Luis and Rodríguez-Miret, Jan and Krallinger, Martin},
    booktitle = {Proceedings of the BioCreative VIII Challenge and Workshop: Curation and Evaluation in the Era of Generative Models},
	month = nov,
	year = {2023},
}

@inproceedings{cormack_reciprocal_2009,
	address = {New York, NY, USA},
	series = {{SIGIR} '09},
	title = {Reciprocal rank fusion outperforms condorcet and individual rank learning methods},
	isbn = {978-1-60558-483-6},
	url = {https://dl.acm.org/doi/10.1145/1571941.1572114},
	doi = {10.1145/1571941.1572114},
	urldate = {2026-04-20},
	booktitle = {Proceedings of the 32nd international {ACM} {SIGIR} conference on {Research} and development in information retrieval},
	author = {Cormack, Gordon V. and Clarke, Charles L A and Buettcher, Stefan},
	month = jul,
	year = {2009},
}

@inproceedings{liu_learning_2021,
	address = {Online},
	title = {Learning {Domain}-{Specialised} {Representations} for {Cross}-{Lingual} {Biomedical} {Entity} {Linking}},
	url = {https://aclanthology.org/2021.acl-short.72/},
	doi = {10.18653/v1/2021.acl-short.72},
	urldate = {2025-12-16},
	booktitle = {Proceedings of the 59th {Annual} {Meeting} of the {Association} for {Computational} {Linguistics} and the 11th {International} {Joint} {Conference} on {Natural} {Language} {Processing} ({Volume} 2: {Short} {Papers})},
	publisher = {Association for Computational Linguistics},
	author = {Liu, Fangyu and Vulić, Ivan and Korhonen, Anna and Collier, Nigel},
	editor = {Zong, Chengqing and Xia, Fei and Li, Wenjie and Navigli, Roberto},
	month = aug,
	year = {2021},
	pages = {565--574},
}

@article{lima-lopez_overview_2023-1,
	title = {Overview of {MedProcNER} task on medical procedure detection and entity linking at {BioASQ} 2023},
	url = {https://ceur-ws.org/Vol-3497/paper-002.pdf},
	journal = {Working Notes of CLEF},
	author = {Lima-López, Salvador and Farré-Maduell, Eulàlia and Gasco, Luis and Nentidis, Anastasios and Krithara, Anastasia and Katsimpras, Georgios and Paliouras, Georgios and Krallinger, Martin},
	year = {2023},
}

@article{bodenreider_unified_2004,
	title = {The {Unified} {Medical} {Language} {System} ({UMLS}): integrating biomedical terminology},
	volume = {32},
	issn = {1362-4962},
	shorttitle = {The {Unified} {Medical} {Language} {System} ({UMLS})},
	doi = {10.1093/nar/gkh061},
	language = {eng},
	number = {Database issue},
	journal = {Nucleic Acids Research},
	author = {Bodenreider, Olivier},
	month = jan,
	year = {2004},
	pmid = {14681409},
	pmcid = {PMC308795},
	pages = {D267--270},
}

@incollection{donnelly_snomed-ct_2006,
	title = {{SNOMED}-{CT}: {The} {Advanced} {Terminology} and {Coding} {System} for {eHealth}},
	shorttitle = {{SNOMED}-{CT}},
	url = {https://ebooks.iospress.nl/publication/9130},
	language = {en},
	urldate = {2026-04-10},
	booktitle = {Medical and {Care} {Compunetics} 3},
	publisher = {IOS Press},
	author = {Donnelly, Kevin},
	year = {2006},
	pages = {279--290},
}

@article{krauthammer_term_2004,
	series = {Named {Entity} {Recognition} in {Biomedicine}},
	title = {Term identification in the biomedical literature},
	volume = {37},
	issn = {1532-0464},
	url = {https://www.sciencedirect.com/science/article/pii/S1532046404000826},
	doi = {10.1016/j.jbi.2004.08.004},
	number = {6},
	urldate = {2026-04-10},
	journal = {Journal of Biomedical Informatics},
	author = {Krauthammer, Michael and Nenadic, Goran},
	month = dec,
	year = {2004},
	pages = {512--526},
}

@inproceedings{wu_scalable_2020,
	address = {Online},
	title = {Scalable {Zero}-shot {Entity} {Linking} with {Dense} {Entity} {Retrieval}},
	url = {https://aclanthology.org/2020.emnlp-main.519/},
	doi = {10.18653/v1/2020.emnlp-main.519},
	urldate = {2025-10-06},
	booktitle = {Proceedings of the 2020 {Conference} on {Empirical} {Methods} in {Natural} {Language} {Processing} ({EMNLP})},
	publisher = {Association for Computational Linguistics},
	author = {Wu, Ledell and Petroni, Fabio and Josifoski, Martin and Riedel, Sebastian and Zettlemoyer, Luke},
	editor = {Webber, Bonnie and Cohn, Trevor and He, Yulan and Liu, Yang},
	month = nov,
	year = {2020},
	pages = {6397--6407},
}

@inproceedings{sakhovskiy_biomedical_2024,
	address = {Mexico City, Mexico},
	title = {Biomedical {Entity} {Representation} with {Graph}-{Augmented} {Multi}-{Objective} {Transformer}},
	url = {https://aclanthology.org/2024.findings-naacl.288/},
	doi = {10.18653/v1/2024.findings-naacl.288},
	urldate = {2026-03-19},
	booktitle = {Findings of the {Association} for {Computational} {Linguistics}: {NAACL} 2024},
	publisher = {Association for Computational Linguistics},
	author = {Sakhovskiy, Andrey and Semenova, Natalia and Kadurin, Artur and Tutubalina, Elena},
	editor = {Duh, Kevin and Gomez, Helena and Bethard, Steven},
	month = jun,
	year = {2024},
	pages = {4626--4643},
}

@article{chen_comprehensive_2026,
	title = {A comprehensive survey on medical concept normalization: {Datasets}, techniques, applications, and future directions},
	volume = {178},
	issn = {1532-0464},
	shorttitle = {A comprehensive survey on medical concept normalization},
	url = {https://www.sciencedirect.com/science/article/pii/S1532046426000298},
	doi = {10.1016/j.jbi.2026.105005},
	urldate = {2026-04-10},
	journal = {Journal of Biomedical Informatics},
	author = {Chen, Haihua and Zhou, Yuhan and Li, Ruochi and Illa, Aryan Murthy and Cleveland, Ana and Ding, Junhua},
	month = jun,
	year = {2026},
	pages = {105005},
}

@incollection{schwartz_simple_2002,
	title = {A simple algorithm for identifying abbreviation definitions in biomedical text},
	isbn = {978-981-238-217-7},
	url = {https://www.worldscientific.com/doi/abs/10.1142/9789812776303_0042},
	doi = {10.1142/9789812776303_0042},
	urldate = {2026-04-10},
	booktitle = {Biocomputing 2003},
	publisher = {WORLD SCIENTIFIC},
	author = {Schwartz, Ariel S. and Hearst, Marti A.},
	month = dec,
	year = {2002},
	pages = {451--462},
}

@inproceedings{liu_self-alignment_2021,
	address = {Online},
	title = {Self-{Alignment} {Pretraining} for {Biomedical} {Entity} {Representations}},
	url = {https://aclanthology.org/2021.naacl-main.334},
	doi = {10.18653/v1/2021.naacl-main.334},
	language = {en},
	urldate = {2024-09-30},
	booktitle = {Proceedings of the 2021 {Conference} of the {North} {American} {Chapter} of the {Association} for {Computational} {Linguistics}: {Human} {Language} {Technologies}},
	publisher = {Association for Computational Linguistics},
	author = {Liu, Fangyu and Shareghi, Ehsan and Meng, Zaiqiao and Basaldella, Marco and Collier, Nigel},
	year = {2021},
	pages = {4228--4238},
}

@inproceedings{agarwal_entity_2022,
	address = {Seattle, United States},
	title = {Entity {Linking} via {Explicit} {Mention}-{Mention} {Coreference} {Modeling}},
	url = {https://aclanthology.org/2022.naacl-main.343/},
	doi = {10.18653/v1/2022.naacl-main.343},
	urldate = {2025-07-17},
	booktitle = {Proceedings of the 2022 {Conference} of the {North} {American} {Chapter} of the {Association} for {Computational} {Linguistics}: {Human} {Language} {Technologies}},
	author = {Agarwal, Dhruv and Angell, Rico and Monath, Nicholas and McCallum, Andrew},
	editor = {Carpuat, Marine and de Marneffe, Marie-Catherine and Meza Ruiz, Ivan Vladimir},
	month = jul,
	year = {2022},
}

@inproceedings{yuan_generative_2022,
	address = {Seattle, United States},
	title = {Generative {Biomedical} {Entity} {Linking} via {Knowledge} {Base}-{Guided} {Pre}-training and {Synonyms}-{Aware} {Fine}-tuning},
	url = {https://aclanthology.org/2022.naacl-main.296/},
	doi = {10.18653/v1/2022.naacl-main.296},
	urldate = {2025-07-17},
	booktitle = {Proceedings of the 2022 {Conference} of the {North} {American} {Chapter} of the {Association} for {Computational} {Linguistics}: {Human} {Language} {Technologies}},
	publisher = {Association for Computational Linguistics},
	author = {Yuan, Hongyi and Yuan, Zheng and Yu, Sheng},
	editor = {Carpuat, Marine and de Marneffe, Marie-Catherine and Meza Ruiz, Ivan Vladimir},
	month = jul,
	year = {2022},
	pages = {4038--4048},
}

@inproceedings{kim_learning_2025,
	address = {Vienna, Austria},
	title = {Learning from {Negative} {Samples} in {Biomedical} {Generative} {Entity} {Linking}},
	isbn = {9798891762565},
	url = {https://aclanthology.org/2025.findings-acl.558/},
	urldate = {2025-07-28},
	booktitle = {Findings of the {Association} for {Computational} {Linguistics}: {ACL} 2025},
	publisher = {Association for Computational Linguistics},
	author = {Kim, Chanhwi and Kim, Hyunjae and Park, Sihyeon and Lee, Jiwoo and Sung, Mujeen and Kang, Jaewoo},
	editor = {Che, Wanxiang and Nabende, Joyce and Shutova, Ekaterina and Pilehvar, Mohammad Taher},
	month = jul,
	year = {2025},
	pages = {10714--10730},
}

@inproceedings{shlyk_real_2024,
	address = {Bangkok, Thailand},
	title = {{REAL}: {A} {Retrieval}-{Augmented} {Entity} {Linking} {Approach} for {Biomedical} {Concept} {Recognition}},
	shorttitle = {{REAL}},
	url = {https://aclanthology.org/2024.bionlp-1.29/},
	doi = {10.18653/v1/2024.bionlp-1.29},
	urldate = {2026-03-19},
	booktitle = {Proceedings of the 23rd {Workshop} on {Biomedical} {Natural} {Language} {Processing}},
	publisher = {Association for Computational Linguistics},
	author = {Shlyk, Darya and Groza, Tudor and Mesiti, Marco and Montanelli, Stefano and Cavalleri, Emanuele},
	editor = {Demner-Fushman, Dina and Ananiadou, Sophia and Miwa, Makoto and Roberts, Kirk and Tsujii, Junichi},
	month = aug,
	year = {2024},
	pages = {380--389},
}

@inproceedings{ratinov_local_2011,
	address = {Portland, Oregon, USA},
	title = {Local and {Global} {Algorithms} for {Disambiguation} to {Wikipedia}},
	url = {https://aclanthology.org/P11-1138/},
	urldate = {2026-04-10},
	booktitle = {Proceedings of the 49th {Annual} {Meeting} of the {Association} for {Computational} {Linguistics}: {Human} {Language} {Technologies}},
	publisher = {Association for Computational Linguistics},
	author = {Ratinov, Lev and Roth, Dan and Downey, Doug and Anderson, Mike},
	editor = {Lin, Dekang and Matsumoto, Yuji and Mihalcea, Rada},
	month = jun,
	year = {2011},
	pages = {1375--1384},
}

@misc{remaki_syncabel_2026,
	title = {{SynCABEL}: {Synthetic} {Contextualized} {Augmentation} for {Biomedical} {Entity} {Linking}},
	shorttitle = {{SynCABEL}},
	url = {http://arxiv.org/abs/2601.19667},
	doi = {10.48550/arXiv.2601.19667},
	urldate = {2026-04-10},
	publisher = {arXiv},
	author = {Remaki, Adam and Gérardin, Christel and Farré-Maduell, Eulàlia and Krallinger, Martin and Tannier, Xavier},
	month = jan,
	year = {2026},
	note = {arXiv:2601.19667 [cs]},
}

@inproceedings{ganea_deep_2017,
	address = {Copenhagen, Denmark},
	title = {Deep {Joint} {Entity} {Disambiguation} with {Local} {Neural} {Attention}},
	url = {https://aclanthology.org/D17-1277/},
	doi = {10.18653/v1/D17-1277},
	urldate = {2026-04-10},
	booktitle = {Proceedings of the 2017 {Conference} on {Empirical} {Methods} in {Natural} {Language} {Processing}},
	publisher = {Association for Computational Linguistics},
	author = {Ganea, Octavian-Eugen and Hofmann, Thomas},
	editor = {Palmer, Martha and Hwa, Rebecca and Riedel, Sebastian},
	month = sep,
	year = {2017},
	pages = {2619--2629},
}

@inproceedings{vollmers_contextual_2025,
	address = {Abu Dhabi, UAE},
	title = {Contextual {Augmentation} for {Entity} {Linking} using {Large} {Language} {Models}},
	url = {https://aclanthology.org/2025.coling-main.570/},
	urldate = {2026-03-19},
	booktitle = {Proceedings of the 31st {International} {Conference} on {Computational} {Linguistics}},
	publisher = {Association for Computational Linguistics},
	author = {Vollmers, Daniel and Zahera, Hamada and Moussallem, Diego and Ngonga Ngomo, Axel-Cyrille},
	editor = {Rambow, Owen and Wanner, Leo and Apidianaki, Marianna and Al-Khalifa, Hend and Eugenio, Barbara Di and Schockaert, Steven},
	month = jan,
	year = {2025},
	pages = {8535--8545},
}

@article{borchert_improving_2024,
	title = {Improving biomedical entity linking for complex entity mentions with {LLM}-based text simplification},
	volume = {2024},
	issn = {1758-0463},
	url = {https://doi.org/10.1093/database/baae067},
	doi = {10.1093/database/baae067},
	urldate = {2026-03-19},
	journal = {Database},
	author = {Borchert, Florian and Llorca, Ignacio and Schapranow, Matthieu-P},
	month = feb,
	year = {2024},
	pages = {baae067},
}

@inproceedings{zhu_llmlink_2025,
	address = {Abu Dhabi, UAE},
	title = {{LlmLink}: {Dual} {LLMs} for {Dynamic} {Entity} {Linking} on {Long} {Narratives} with {Collaborative} {Memorisation} and {Prompt} {Optimisation}},
	shorttitle = {{LlmLink}},
	url = {https://aclanthology.org/2025.coling-main.751/},
	urldate = {2026-04-10},
	booktitle = {Proceedings of the 31st {International} {Conference} on {Computational} {Linguistics}},
	publisher = {Association for Computational Linguistics},
	author = {Zhu, Lixing and Wang, Jun and He, Yulan},
	editor = {Rambow, Owen and Wanner, Leo and Apidianaki, Marianna and Al-Khalifa, Hend and Eugenio, Barbara Di and Schockaert, Steven},
	month = jan,
	year = {2025},
	pages = {11334--11347},
}

@inproceedings{neumann_scispacy_2019,
	address = {Florence, Italy},
	title = {{ScispaCy}: {Fast} and {Robust} {Models} for {Biomedical} {Natural} {Language} {Processing}},
	shorttitle = {{ScispaCy}},
	url = {https://aclanthology.org/W19-5034/},
	doi = {10.18653/v1/W19-5034},
	urldate = {2025-07-17},
	booktitle = {Proceedings of the 18th {BioNLP} {Workshop} and {Shared} {Task}},
	publisher = {Association for Computational Linguistics},
	author = {Neumann, Mark and King, Daniel and Beltagy, Iz and Ammar, Waleed},
	editor = {Demner-Fushman, Dina and Cohen, Kevin Bretonnel and Ananiadou, Sophia and Tsujii, Junichi},
	month = aug,
	year = {2019},
	pages = {319--327},
}

@article{aronson_overview_2010,
	title = {An overview of {MetaMap}: historical perspective and recent advances},
	volume = {17},
	issn = {1067-5027},
	shorttitle = {An overview of {MetaMap}},
	url = {https://doi.org/10.1136/jamia.2009.002733},
	doi = {10.1136/jamia.2009.002733},
	number = {3},
	urldate = {2025-01-07},
	journal = {Journal of the American Medical Informatics Association},
	author = {Aronson, Alan R and Lang, François-Michel},
	month = may,
	year = {2010},
	pages = {229--236},
}

@article{savova_mayo_2010,
	title = {Mayo clinical {Text} {Analysis} and {Knowledge} {Extraction} {System} ({cTAKES}): architecture, component evaluation and applications},
	volume = {17},
	issn = {1067-5027},
	shorttitle = {Mayo clinical {Text} {Analysis} and {Knowledge} {Extraction} {System} ({cTAKES})},
	url = {https://doi.org/10.1136/jamia.2009.001560},
	doi = {10.1136/jamia.2009.001560},
	number = {5},
	urldate = {2025-01-07},
	journal = {Journal of the American Medical Informatics Association},
	author = {Savova, Guergana K and Masanz, James J and Ogren, Philip V and Zheng, Jiaping and Sohn, Sunghwan and Kipper-Schuler, Karin C and Chute, Christopher G},
	month = sep,
	year = {2010},
	pages = {507--513},
}

@misc{grattafiori_llama_2024,
	title = {The {Llama} 3 {Herd} of {Models}},
	url = {http://arxiv.org/abs/2407.21783},
	doi = {10.48550/arXiv.2407.21783},
	urldate = {2025-07-18},
	publisher = {arXiv},
    author={Grattafiori, Aaron and Dubey, Abhimanyu and Jauhri, Abhinav and Pandey, Abhinav and others},
	month = nov,
	year = {2024},
	note = {arXiv:2407.21783 [cs]},
}

@misc{mohan_medmentions_2019,
	title = {{MedMentions}: {A} {Large} {Biomedical} {Corpus} {Annotated} with {UMLS} {Concepts}},
	shorttitle = {{MedMentions}},
	url = {http://arxiv.org/abs/1902.09476},
	doi = {10.48550/arXiv.1902.09476},
	urldate = {2025-07-17},
	publisher = {arXiv},
	author = {Mohan, Sunil and Li, Donghui},
	month = feb,
	year = {2019},
	note = {arXiv:1902.09476 [cs]},
}

@inproceedings{wang_aelc_2025,
	address = {Suzhou, China},
	title = {{AELC}: {Adaptive} {Entity} {Linking} with {LLM}-{Driven} {Contextualization}},
	isbn = {979-8-89176-335-7},
	shorttitle = {{AELC}},
	url = {https://aclanthology.org/2025.findings-emnlp.231/},
	doi = {10.18653/v1/2025.findings-emnlp.231},
	urldate = {2026-03-19},
	booktitle = {Findings of the {Association} for {Computational} {Linguistics}: {EMNLP} 2025},
	publisher = {Association for Computational Linguistics},
	author = {Wang, Fang and Tao, Zhengwei and Wang, Ming and Hu, Minghao and Bai, Xiaoying},
	editor = {Christodoulopoulos, Christos and Chakraborty, Tanmoy and Rose, Carolyn and Peng, Violet},
	month = nov,
	year = {2025},
	pages = {4313--4327},
}

@inproceedings{ding_chatel_2024,
	address = {Torino, Italia},
	title = {{ChatEL}: {Entity} {Linking} with {Chatbots}},
	shorttitle = {{ChatEL}},
	url = {https://aclanthology.org/2024.lrec-main.275/},
	urldate = {2026-03-19},
	booktitle = {Proceedings of the 2024 {Joint} {International} {Conference} on {Computational} {Linguistics}, {Language} {Resources} and {Evaluation} ({LREC}-{COLING} 2024)},
	author = {Ding, Yifan and Zeng, Qingkai and Weninger, Tim},
	editor = {Calzolari, Nicoletta and Kan, Min-Yen and Hoste, Veronique and Lenci, Alessandro and Sakti, Sakriani and Xue, Nianwen},
	month = may,
	year = {2024},
}

@inproceedings{xin_llmael_2025,
	address = {New York, NY, USA},
	series = {{CIKM} '25},
	title = {{LLMAEL}: {Large} {Language} {Models} are {Good} {Context} {Augmenters} for {Entity} {Linking}},
	isbn = {979-8-4007-2040-6},
	url = {https://doi.org/10.1145/3746252.3761156},
	doi = {10.1145/3746252.3761156},
	booktitle = {Proceedings of the 34th {ACM} {International} {Conference} on {Information} and {Knowledge} {Management}},
	publisher = {Association for Computing Machinery},
	author = {Xin, Amy and Qi, Yunjia and Yao, Zijun and Zhu, Fangwei and Zeng, Kaisheng and Xu, Bin and Hou, Lei and Li, Juanzi},
	year = {2025},
	pages = {3550--3559},
}

@inproceedings{lin_guiding_2025,
	address = {Suzhou, China},
	title = {Guiding {Large} {Language} {Models} for {Biomedical} {Entity} {Linking} via {Restrictive} and {Contrastive} {Decoding}},
	isbn = {979-8-89176-335-7},
	url = {https://aclanthology.org/2025.findings-emnlp.1292/},
	doi = {10.18653/v1/2025.findings-emnlp.1292},
	urldate = {2026-03-19},
	booktitle = {Findings of the {Association} for {Computational} {Linguistics}: {EMNLP} 2025},
	publisher = {Association for Computational Linguistics},
	author = {Lin, Zhenxi and Zhang, Ziheng and Wu, Jian and Zheng, Yefeng and Wu, Xian},
	editor = {Christodoulopoulos, Christos and Chakraborty, Tanmoy and Rose, Carolyn and Peng, Violet},
	month = nov,
	year = {2025},
	pages = {23745--23759},
}

@inproceedings{cao_autoregressive_2021,
	title = {Autoregressive {Entity} {Retrieval}},
	url = {https://openreview.net/forum?id=5k8F6UU39V},
	language = {en},
	urldate = {2026-04-10},
	author = {Cao, Nicola De and Izacard, Gautier and Riedel, Sebastian and Petroni, Fabio},
	month = oct,
	year = {2021},
    booktitle = {International Conference on Learning Representations},
}

@article{yuan_coder_2022,
	title = {{CODER}: {Knowledge}-infused cross-lingual medical term embedding for term normalization},
	volume = {126},
	issn = {1532-0464},
	shorttitle = {{CODER}},
	url = {https://www.sciencedirect.com/science/article/pii/S1532046421003129},
	doi = {10.1016/j.jbi.2021.103983},
	urldate = {2024-03-28},
	journal = {Journal of Biomedical Informatics},
	author = {Yuan, Zheng and Zhao, Zhengyun and Sun, Haixia and Li, Jiao and Wang, Fei and Yu, Sheng},
	month = feb,
	year = {2022},
	pages = {103983},
}

\appendix

\section{Robust Memory Illustration}
\label{sec:appendix_robust_memory}

\begin{figure*}
    \centering
    \includegraphics[width=0.8\textwidth]{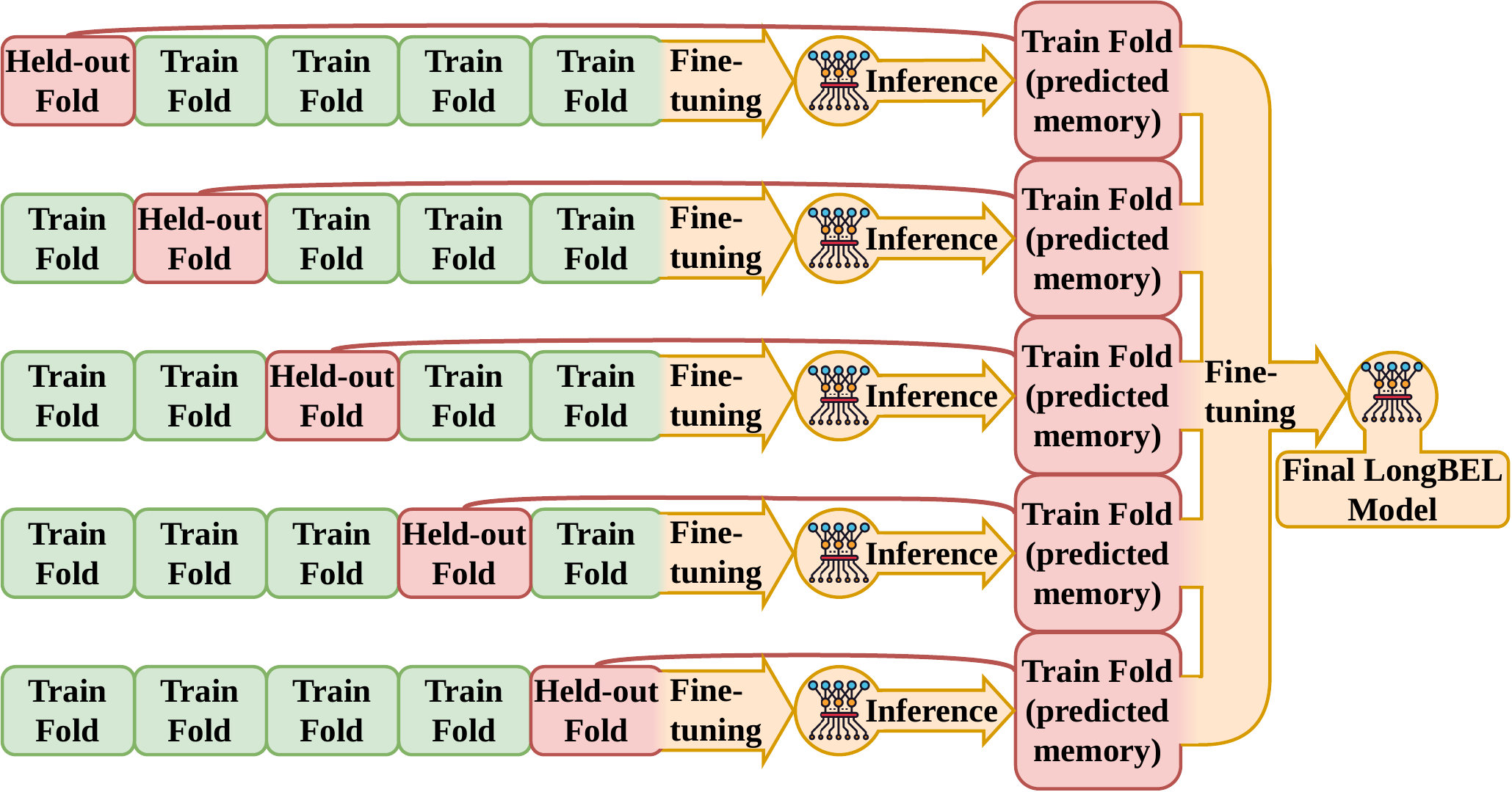}
    \caption{
    Illustration of robust memory construction via cross-validation. The training set is split into five folds. For each fold, a base model is trained on the remaining folds and used to generate predictions on the held-out fold. These held-out predictions are then used to create the memory during final LongBEL training, exposing the model to realistic inference-time memory.
    }
    \label{fig:robust_memory}
\end{figure*}

Figure~\ref{fig:robust_memory} illustrates how cross-validated predictions are used to construct realistic memory for robust LongBEL training.

\section{Ensemble Inference Illustration}
\label{app:ensemble_inference}
\begin{figure*}
    \centering
    \includegraphics[width=0.95\textwidth]{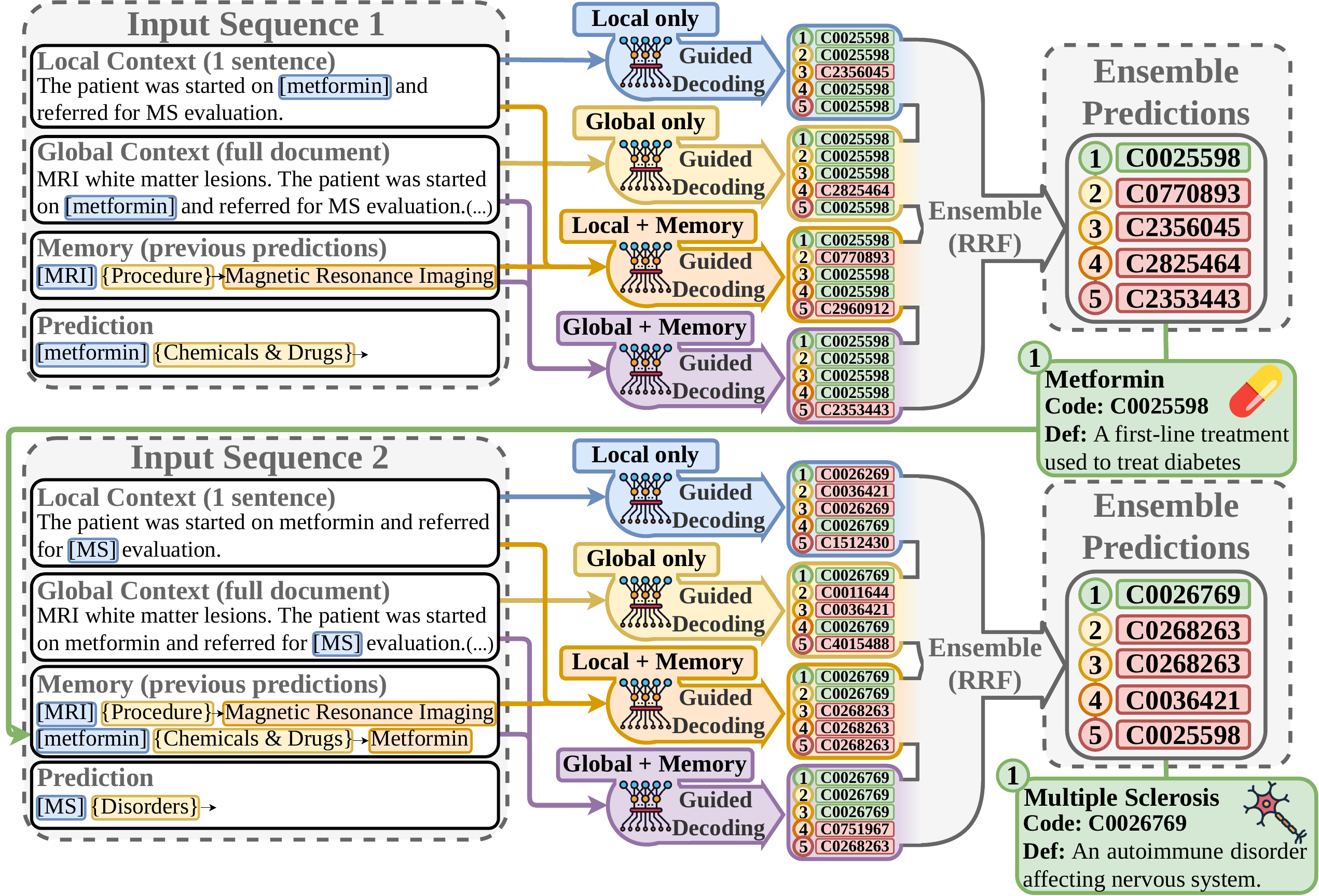}
    \caption{Overview of LongBEL-Ensemble inference. For each mention, LongBEL-Ensemble combines local context, full-document context, and memory of previous predictions to generate multiple ranked candidate lists using semantic-guided constrained decoding. These lists are merged with Reciprocal Rank Fusion (RRF) to produce the final prediction. The predicted concept is then added to memory and used for subsequent mentions.}
    \label{fig:ensemble_inference}
\end{figure*}

Figure~\ref{fig:ensemble_inference} illustrates the complete LongBEL ensemble procedure. For each mention, four input configurations are considered: local context only, global context only, local context with memory, and global context with memory. Each configuration produces a ranked list of candidate concepts using constrained beam search. The four lists are then combined using Reciprocal Rank Fusion, and the highest-ranked concept is selected. The selected prediction is added to memory before the next mention is processed.

\section{Dataset Statistics}
\label{sec:dataset_statistics}

Table~\ref{tab:dataset-summary} reports the size, document characteristics, semantic groups, and concept recurrence of each evaluation dataset.

\begin{table}[h]
\centering
\small
\setlength{\tabcolsep}{1.5pt}
\renewcommand{\arraystretch}{0.9}
\begin{tabular}{@{}p{1.7cm}
                >{\centering\arraybackslash}p{1.0cm}
                >{\centering\arraybackslash}p{1.2cm}
                >{\centering\arraybackslash}p{1.1cm}
                >{\centering\arraybackslash}p{1.1cm}
                >{\centering\arraybackslash}p{1.1cm}@{}}
\toprule
 & MM-ST21pv
 & QUAERO-EMEA
 & \shortstack{Symp\\TEMIST}
 & \shortstack{Dis\\TEMIST}
 & \shortstack{MedProc\\NER} \\
\midrule
\# Docs       & 4,392   & 26     & 1,000  & 1,000  & 1,000  \\
Avg. Len.     & 1,581   & 5,062  & 2,337  & 2,337  & 2,337  \\
Language      & En.     & Fr.    & Es.    & Es.    & Es.    \\
Clinical      & \xmark  & \xmark & \cmark & \cmark & \cmark \\
\# Mentions   & 203,185 & 7,127  & 11,683 & 7,041  & 8,379  \\
\# Concepts   & 25,419  & 1,226  & 4,161  & 3,015  & 2,351  \\
\# Sem. Grps  & 14      & 10     & 1      & 1      & 1      \\
\% First      & 48.5\%  & 41.2\% & 88.8\% & 84.6\% & 83.7\% \\
\% Recurring  & 51.5\%  & 58.8\% & 11.2\% & 15.4\% & 16.3\% \\
\bottomrule
\end{tabular}
\caption{Dataset statistics. Avg. Len.: average document length in characters; Clinical: whether the documents are clinical reports; \# Sem. Grps: number of semantic groups; \% First and \% Recurring: mentions whose concept has not and has previously appeared in the document, respectively.}
\label{tab:dataset-summary}
\end{table}

\section{Knowledge Bases}
\label{sec:appendix_kb}

For each corpus, we selected a KB consistent with its original annotation setup. For MM‑ST21pv, we use UMLS 2017AA, narrowed down to the ST21‑pv subset which covers 21 semantic types with synonyms sourced from 18 vocabularies. QUAERO is paired with UMLS 2014AA, from which we retained only the 10 semantic groups relevant to the corpus. For the three SPACCC datasets, candidate concepts were drawn from gazetteers built from the appropriate branches of the Spanish edition of SNOMED-CT (July 31, 2021 release). The statistics of these KB subsets are summarized in Table \ref{tab:kb_stats}.

\begin{table}[ht]
\centering
\resizebox{\columnwidth}{!}{
\begin{tabular}{@{}lccc@{}}
\toprule
                        & \textbf{MM-ST21pv} & \textbf{QUAERO} & \textbf{SPACCC} \\
\midrule
Version & 2017AA            & 2014AA          & 2021 \\
\# Concepts             & 2,374,392         & 2,960,365       & 276,670 \\
\# Synonyms             & 4,905,087         & 6,970,593       & 546,814 \\
\# Semantic Groups      & 14                & 10              & 3 \\
\# Synonyms per Code    & 2.1               & 2.4             & 1.5 \\
\bottomrule
\end{tabular}
}
\caption{Summary statistics of the knowledge base subsets used for candidate retrieval across the evaluated datasets.}
\label{tab:kb_stats}
\end{table}

\section{Training and Inference Setup}
\label{sec:appendix_implementation}

We train our models using the \texttt{trl} \texttt{SFTTrainer} library. All detailed hyperparameters are provided in Tables~\ref{tab:train_hyperparams} and~\ref{tab:infer_hyperparams}.

\begin{table}[ht]
\centering
\resizebox{\columnwidth}{!}{
\begin{tabular}{@{}ll@{}}
\toprule
\textbf{Hyperparameter} & \textbf{Value} \\
\midrule
Learning Rate & $3 \times 10^{-5}$ \\
Learning Rate Scheduler & Linear \\
Warmup Ratio & $0.03$ \\
Epochs & $50$ \\
Precision & \texttt{BF16} \\
Attention Implementation & Flash Attention 2 \\
Gradient Accumulation Steps & $1$ \\
Dataset Max Input Length & $L_{\max}$ tokens \\
Token Budget per Batch & $16{,}384$ tokens \\
Per-device Batch Size & $\left\lfloor 16{,}384 / L_{\max} \right\rfloor$ \\
\bottomrule
\end{tabular}
}
\caption{Supervised Fine-Tuning (SFT) hyperparameters used across all evaluated models. $L_{\max}$ is the maximum tokenized input length over the training split.}
\label{tab:train_hyperparams}
\end{table}

\begin{table}[ht]
\centering
\begin{tabular}{@{}ll@{}}
\toprule
\textbf{Hyperparameter} & \textbf{Value} \\
\midrule
Decoding Strategy & Beam Search \\
Number of Beams & 5 \\
Batch Size & 16 \\
Precision & \texttt{BF16} \\
\bottomrule
\end{tabular}
\caption{Inference hyperparameters used for the generation phase.}
\label{tab:infer_hyperparams}
\end{table}

\section{Fine-grained Analysis by Semantic Group}
\label{sec:appendix_sem_group}

\begin{figure}[t]
    \centering
    \includegraphics[width=0.5\textwidth]{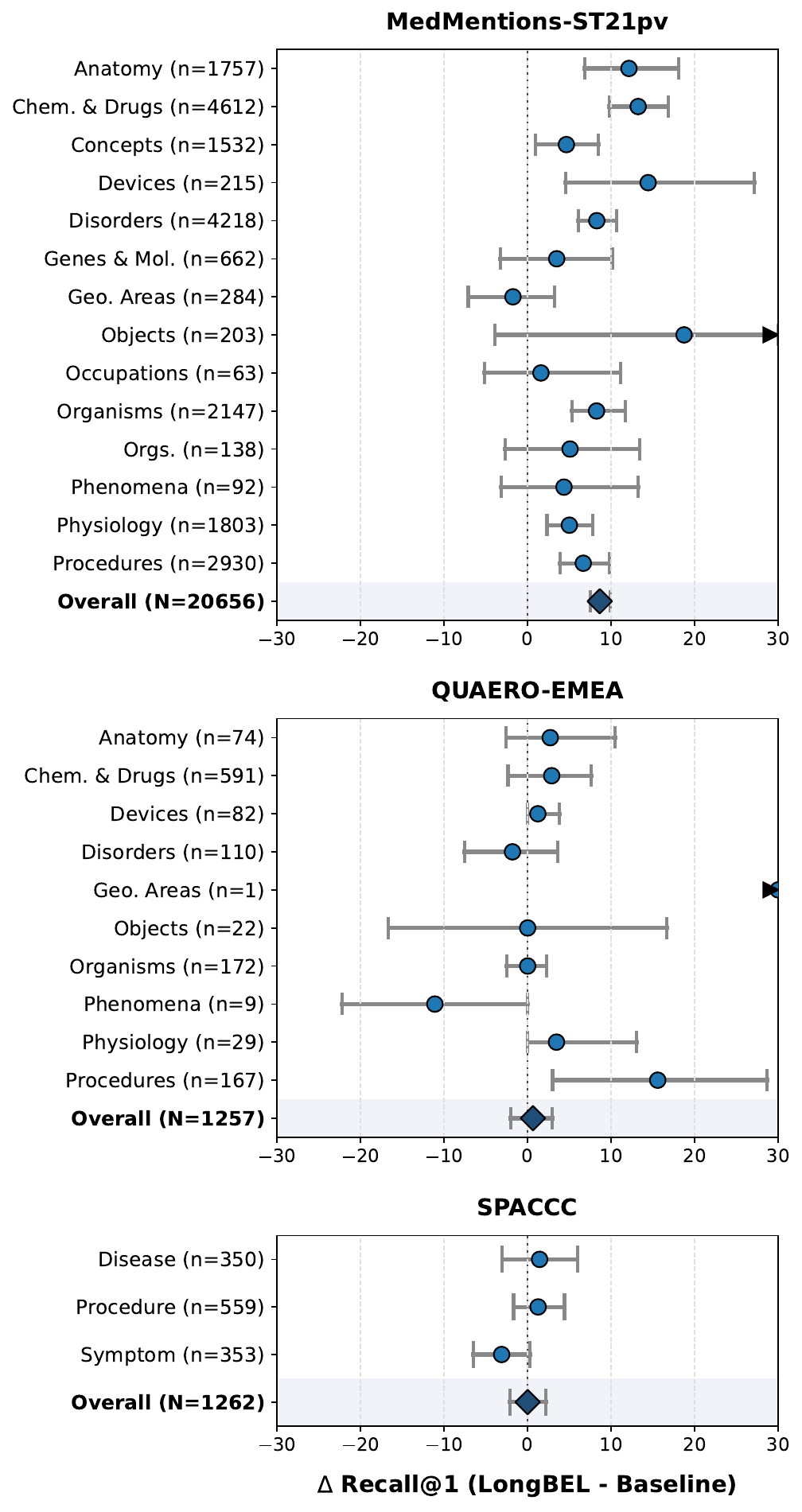}
    \caption{Difference in Recall@1 between LongBEL and the sentence-level baseline, computed on mentions of repeated concepts only, for the LongBEL-1B model. Results are reported per semantic group across datasets.}
    \label{fig:delta_recall_1b}
\end{figure}

\begin{figure}[t]
    \centering
    \includegraphics[width=0.5\textwidth]{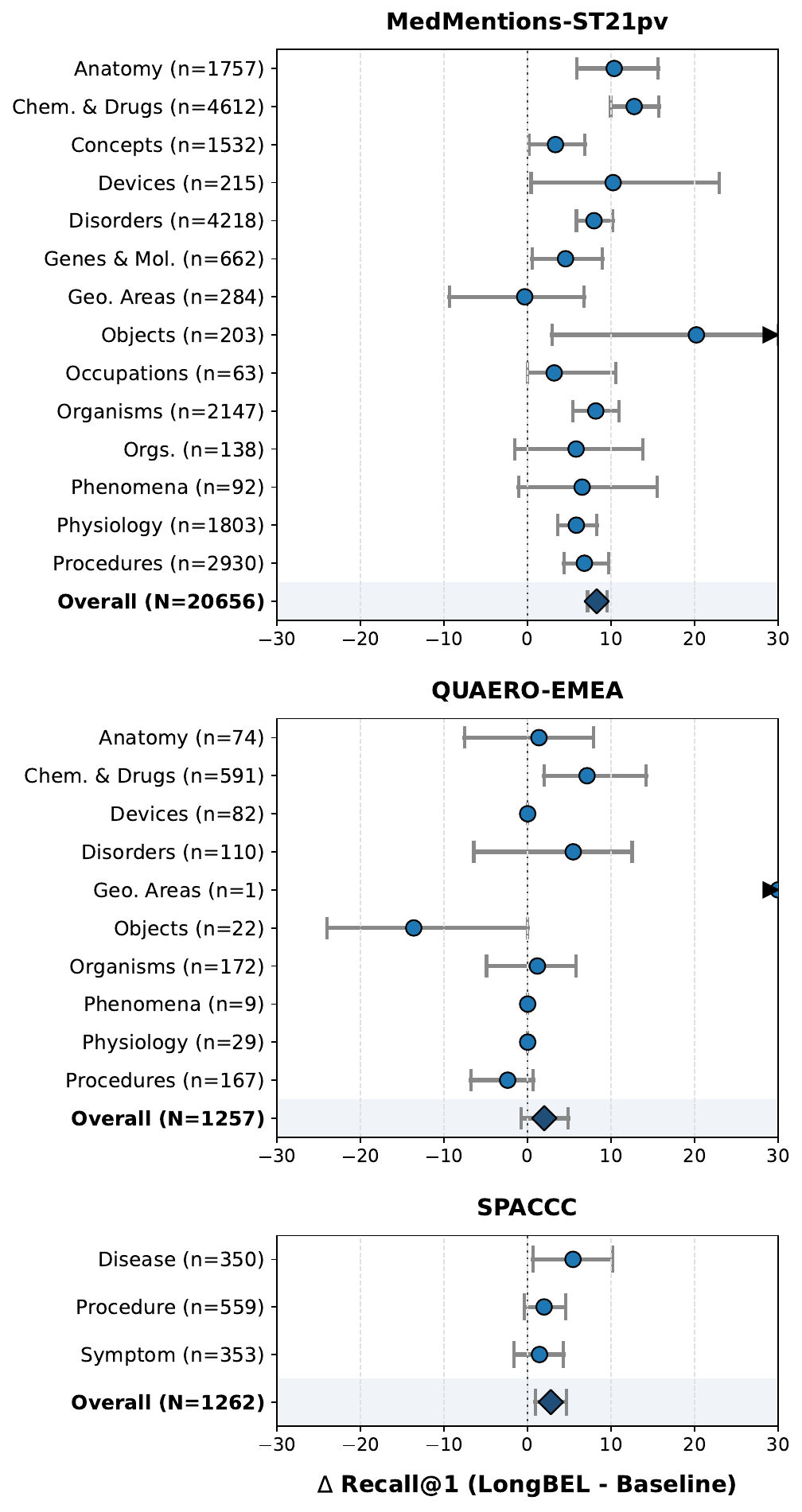}
    \caption{Difference in Recall@1 between LongBEL and the sentence-level baseline, computed on mentions of repeated concepts only, for the LongBEL-8B model. Results are reported per semantic group across datasets.}
    \label{fig:delta_recall_8b}
\end{figure}

We report $\Delta$ Recall, defined as the Recall@1 difference between LongBEL and the sentence-level baseline. Figures~\ref{fig:delta_recall_1b} and~\ref{fig:delta_recall_8b} show the results for LongBEL-1B and LongBEL-8B by semantic group and dataset. LongBEL improves recall across most semantic groups, showing that its benefits are not limited to one type of entity. On MM-ST21pv, gains are relatively stable across groups, with larger improvements for frequent and ambiguous groups such as \textit{Chemicals}, \textit{Disorders}, and \textit{Procedures}. Smaller semantic groups show more variation, likely because they contain fewer examples. On SPACCC and QUAERO-EMEA, the gains are generally smaller and more variable across groups. This is likely due to both lower concept redundancy and smaller sample sizes in some categories, which make group-level estimates less stable. Overall, these results show that LongBEL provides broad improvements across semantic groups, while the size and stability of the gains depend on both concept redundancy and sample size.

\section{Detailed Subset Results}
\label{sec:appendix_detailed_subset_results}

Tables~\ref{tab:detailed_mm_st21pv}--\ref{tab:detailed_medprocner} report detailed subset results for the sentence-level baselines, LongBEL, and the ensemble. The first two rows report overall Recall@1 and Recall@5. The other rows report Recall@1 on different subsets. The support column gives the percentage and number of examples in each subset. Each score is reported with half of the 95\% confidence interval.

The subsets capture simple properties of the mentions. Seen and unseen concepts indicate whether the gold concept appears in the training set. Identical and non-identical mentions indicate whether the mention text exactly matches a synonym of the gold concept in the knowledge base. We also group mentions by length and by document recurrence. First occurrence means that the gold concept has not appeared earlier in the document, while recurring occurrence means that it has already appeared before. For the Spanish SPACCC datasets only, we also separate simple and composite concepts. Simple concepts map to one concept, while composite concepts map to a combination of several concepts.

These tables add detail to the main analysis. MM-ST21pv is much larger than QUAERO-EMEA and SPACCC, so its subset trends are more stable. In MM-ST21pv, LongBEL changes little on first occurrences but improves clearly on recurring mentions, with the largest gains on fifth-or-later occurrences. The smaller datasets show less stable trends because some subsets have few examples and large confidence intervals.

The tables also highlight dataset differences. The SPACCC datasets contain many more multi-word mentions than MM-ST21pv and QUAERO-EMEA, which makes the normalization setting harder. They also include composite concepts, which are specific to these datasets and remain difficult for all models. In SPACCC, a mention may be linked to multiple concepts rather than a single concept. To support this setting, we extend constrained decoding with a dedicated separator token: ‘‘\texttt{<+>}''. During generation, the model predicts concept names as usual and emits the separator token ‘‘\texttt{<+>}'' whenever an additional concept is required, allowing multiple KB concepts to be generated in a single output sequence while preserving constrained decoding.

\begin{table*}
\centering
\small
\setlength{\tabcolsep}{1.4pt}
\renewcommand{\arraystretch}{1.1}
\begin{tabularx}{\textwidth}{>{\raggedright\arraybackslash}p{0.18\textwidth}rcccccc}
\toprule
\textbf{Subset} & \textbf{Support} & \textbf{Baseline-1B} & \textbf{LongBEL-1B} & \textbf{Ensemble-1B} & \textbf{Baseline-8B} & \textbf{LongBEL-8B} & \textbf{Ensemble-8B} \\
\midrule
\multicolumn{8}{l}{\textbf{Overall performance}} \\
Recall@1 overall & 100\% (40,143) & 73.1$_{\pm1.0}$ & 77.6$_{\pm0.9}$ & 79.1$_{\pm0.8}$ & 75.0$_{\pm0.9}$ & 79.3$_{\pm0.8}$ & 80.6$_{\pm0.8}$ \\
Recall@5 overall & 100\% (40,143) & 81.2$_{\pm0.8}$ & 85.8$_{\pm0.7}$ & 84.1$_{\pm0.8}$ & 83.7$_{\pm0.9}$ & 87.2$_{\pm0.8}$ & 85.6$_{\pm0.7}$ \\
\midrule
\multicolumn{8}{l}{\textbf{Concept familiarity}} \\
Seen concepts & 79.5\% (31,914) & 80.5$_{\pm0.9}$ & 82.8$_{\pm0.8}$ & 84.0$_{\pm0.8}$ & 82.2$_{\pm0.8}$ & 84.6$_{\pm0.8}$ & 85.3$_{\pm0.7}$ \\
Unseen concepts & 20.5\% (8,229) & 44.6$_{\pm2.1}$ & 57.5$_{\pm2.4}$ & 57.7$_{\pm2.4}$ & 47.1$_{\pm2.2}$ & 59.1$_{\pm2.5}$ & 59.6$_{\pm2.5}$ \\
\midrule
\multicolumn{8}{l}{\textbf{Lexical match}} \\
Identical & 5.6\% (2,265) & 93.5$_{\pm1.9}$ & 95.5$_{\pm1.3}$ & 96.0$_{\pm1.3}$ & 94.6$_{\pm1.4}$ & 94.3$_{\pm1.4}$ & 95.5$_{\pm1.5}$ \\
Not identical & 94.4\% (37,878) & 71.9$_{\pm0.9}$ & 76.5$_{\pm0.9}$ & 77.6$_{\pm0.9}$ & 73.8$_{\pm0.9}$ & 78.4$_{\pm0.8}$ & 79.1$_{\pm0.9}$ \\
\midrule
\multicolumn{8}{l}{\textbf{Mention length}} \\
Single word & 66.8\% (26,833) & 74.9$_{\pm1.1}$ & 80.4$_{\pm1.0}$ & 81.4$_{\pm1.0}$ & 76.6$_{\pm1.1}$ & 82.3$_{\pm1.0}$ & 82.6$_{\pm0.9}$ \\
Multi-word & 33.2\% (13,310) & 69.5$_{\pm1.2}$ & 72.0$_{\pm1.2}$ & 73.0$_{\pm1.1}$ & 71.7$_{\pm1.1}$ & 73.5$_{\pm1.1}$ & 74.7$_{\pm1.1}$ \\
% $>$ 3 words & 2.9\% (1,147) & 56.1$_{\pm3.5}$ & 60.5$_{\pm3.4}$ & 59.6$_{\pm3.6}$ & 61.0$_{\pm3.3}$ & 63.2$_{\pm3.3}$ & 63.6$_{\pm3.3}$ \\
\midrule
\multicolumn{8}{l}{\textbf{Document recurrence}} \\
First occurrence & 48.5\% (19,487) & 79.1$_{\pm0.7}$ & 79.1$_{\pm0.7}$ & 80.6$_{\pm0.6}$ & 80.2$_{\pm0.6}$ & 80.3$_{\pm0.7}$ & 81.7$_{\pm0.6}$ \\
Recurring occurrence & 51.5\% (20,656) & 67.5$_{\pm1.4}$ & 76.1$_{\pm1.4}$ & 76.8$_{\pm1.3}$ & 70.1$_{\pm1.5}$ & 78.4$_{\pm1.3}$ & 78.4$_{\pm1.2}$ \\
2nd occurrence & 18.5\% (7,436) & 75.8$_{\pm1.1}$ & 78.6$_{\pm1.0}$ & 79.8$_{\pm1.0}$ & 77.1$_{\pm1.0}$ & 80.4$_{\pm0.9}$ & 80.8$_{\pm0.9}$ \\
3rd occurrence & 10.4\% (4,167) & 69.6$_{\pm1.6}$ & 77.1$_{\pm1.4}$ & 77.8$_{\pm1.3}$ & 72.2$_{\pm1.6}$ & 79.2$_{\pm1.4}$ & 79.0$_{\pm1.3}$ \\
4th occurrence & 6.7\% (2,704) & 65.0$_{\pm1.9}$ & 75.8$_{\pm1.7}$ & 76.3$_{\pm1.7}$ & 68.2$_{\pm1.8}$ & 77.8$_{\pm1.5}$ & 78.1$_{\pm1.5}$ \\
5th+ occurrence & 15.8\% (6,349) & 57.3$_{\pm2.8}$ & 72.7$_{\pm2.5}$ & 72.8$_{\pm2.5}$ & 61.4$_{\pm2.6}$ & 75.9$_{\pm2.4}$ & 75.4$_{\pm2.4}$ \\
\bottomrule
\end{tabularx}

\caption{Detailed subset results on MM-ST21pv. The first two rows report overall Recall@1 and Recall@5. All other rows report Recall@1 on specific mention subsets. The support column gives the percentage and number of examples in each subset. Each score is shown with half of the 95\% confidence interval. First occurrence refers to mentions whose gold concept has not appeared earlier in the document, while recurring occurrence refers to mentions whose gold concept has already appeared earlier.}
\label{tab:detailed_mm_st21pv}
\end{table*}

\begin{table*}
\centering
\small
\setlength{\tabcolsep}{1.4pt}
\renewcommand{\arraystretch}{1.1}
\begin{tabularx}{\textwidth}{>{\raggedright\arraybackslash}p{0.18\textwidth}rcccccc}
\toprule
\textbf{Subset} & \textbf{Support} & \textbf{Baseline-1B} & \textbf{LongBEL-1B} & \textbf{Ensemble-1B} & \textbf{Baseline-8B} & \textbf{LongBEL-8B} & \textbf{Ensemble-8B} \\
\midrule
\multicolumn{8}{l}{\textbf{Overall performance}} \\
Recall@1 overall & 100\% (2,196) & 75.1$_{\pm3.6}$ & 74.5$_{\pm3.7}$ & 77.2$_{\pm3.0}$ & 73.8$_{\pm4.0}$ & 75.4$_{\pm3.4}$ & 77.6$_{\pm3.0}$ \\
Recall@5 overall & 100\% (2,196) & 82.0$_{\pm3.0}$ & 82.4$_{\pm2.8}$ & 82.8$_{\pm3.2}$ & 82.2$_{\pm2.6}$ & 82.0$_{\pm2.5}$ & 82.8$_{\pm2.7}$ \\
\midrule
\multicolumn{8}{l}{\textbf{Concept familiarity}} \\
Seen concepts & 56.0\% (1,229) & 88.7$_{\pm2.7}$ & 89.3$_{\pm2.1}$ & 91.1$_{\pm1.8}$ & 92.1$_{\pm2.1}$ & 90.8$_{\pm2.8}$ & 92.8$_{\pm1.2}$ \\
Unseen concepts & 44.0\% (967) & 57.5$_{\pm5.8}$ & 55.5$_{\pm5.5}$ & 59.2$_{\pm5.6}$ & 50.2$_{\pm5.1}$ & 55.8$_{\pm5.4}$ & 58.0$_{\pm5.6}$ \\
\midrule
\multicolumn{8}{l}{\textbf{Lexical match}} \\
Identical & 6.5\% (143) & 93.7$_{\pm4.4}$ & 97.9$_{\pm2.5}$ & 100.0$_{\pm0.0}$ & 93.0$_{\pm4.6}$ & 99.3$_{\pm1.0}$ & 98.6$_{\pm1.8}$ \\
Not identical & 93.5\% (2,053) & 73.8$_{\pm3.4}$ & 72.8$_{\pm3.9}$ & 75.6$_{\pm3.1}$ & 72.5$_{\pm4.0}$ & 73.7$_{\pm3.6}$ & 76.2$_{\pm2.9}$ \\
\midrule
\multicolumn{8}{l}{\textbf{Mention length}} \\
Single word & 77.6\% (1,704) & 79.9$_{\pm4.0}$ & 80.8$_{\pm4.0}$ & 82.1$_{\pm3.5}$ & 78.5$_{\pm4.8}$ & 80.3$_{\pm4.9}$ & 82.0$_{\pm3.7}$ \\
Multi-word & 22.4\% (492) & 58.5$_{\pm6.4}$ & 52.6$_{\pm9.1}$ & 60.2$_{\pm8.2}$ & 57.7$_{\pm5.7}$ & 58.3$_{\pm6.0}$ & 62.4$_{\pm5.6}$ \\
% $>$ 3 words & 3.4\% (74) & 64.9$_{\pm14.4}$ & 51.4$_{\pm9.6}$ & 58.1$_{\pm7.9}$ & 60.8$_{\pm5.7}$ & 58.1$_{\pm11.1}$ & 64.9$_{\pm10.1}$ \\
\midrule
\multicolumn{8}{l}{\textbf{Document recurrence}} \\
First occurrence & 42.8\% (939) & 75.0$_{\pm3.4}$ & 72.6$_{\pm3.6}$ & 75.6$_{\pm3.5}$ & 73.6$_{\pm3.1}$ & 74.2$_{\pm3.5}$ & 76.3$_{\pm2.8}$ \\
Recurring occurrence & 57.2\% (1,257) & 75.2$_{\pm5.4}$ & 75.8$_{\pm5.1}$ & 78.4$_{\pm4.6}$ & 74.0$_{\pm5.5}$ & 76.3$_{\pm4.9}$ & 78.7$_{\pm4.3}$ \\
2nd occurrence & 18.9\% (416) & 77.4$_{\pm2.7}$ & 76.0$_{\pm4.6}$ & 79.1$_{\pm3.3}$ & 76.2$_{\pm4.4}$ & 77.6$_{\pm3.2}$ & 79.3$_{\pm3.2}$ \\
3rd occurrence & 9.6\% (210) & 72.4$_{\pm5.7}$ & 71.4$_{\pm5.4}$ & 75.7$_{\pm4.1}$ & 72.9$_{\pm4.4}$ & 74.8$_{\pm3.5}$ & 75.2$_{\pm4.2}$ \\
4th occurrence & 5.5\% (120) & 77.5$_{\pm5.3}$ & 76.7$_{\pm5.0}$ & 78.3$_{\pm4.8}$ & 75.8$_{\pm6.0}$ & 76.7$_{\pm5.5}$ & 79.2$_{\pm4.4}$ \\
5th+ occurrence & 23.3\% (511) & 74.0$_{\pm9.6}$ & 77.3$_{\pm9.2}$ & 78.9$_{\pm8.6}$ & 72.2$_{\pm10.1}$ & 75.7$_{\pm9.7}$ & 79.5$_{\pm8.2}$ \\
\bottomrule
\end{tabularx}

\caption{Detailed subset results on QUAERO-EMEA. The first two rows report overall Recall@1 and Recall@5. All other rows report Recall@1 on specific mention subsets. The support column gives the percentage and number of examples in each subset. Each score is shown with half of the 95\% confidence interval. First occurrence refers to mentions whose gold concept has not appeared earlier in the document, while recurring occurrence refers to mentions whose gold concept has already appeared earlier.}
\label{tab:detailed_quaero_emea}
\end{table*}

\begin{table*}
\centering
\small
\setlength{\tabcolsep}{1.4pt}
\renewcommand{\arraystretch}{1.1}
\begin{tabularx}{\textwidth}{>{\raggedright\arraybackslash}p{0.18\textwidth}rcccccc}
\toprule
\textbf{Subset} & \textbf{Support} & \textbf{Baseline-1B} & \textbf{LongBEL-1B} & \textbf{Ensemble-1B} & \textbf{Baseline-8B} & \textbf{LongBEL-8B} & \textbf{Ensemble-8B} \\
\midrule
\multicolumn{8}{l}{\textbf{Overall performance}} \\
Recall@1 overall & 100\% (2,898) & 60.5$_{\pm2.4}$ & 59.8$_{\pm2.5}$ & 61.8$_{\pm2.5}$ & 61.7$_{\pm2.5}$ & 62.0$_{\pm2.6}$ & 63.3$_{\pm2.5}$ \\
Recall@5 overall & 100\% (2,898) & 68.2$_{\pm2.4}$ & 66.1$_{\pm2.4}$ & 70.5$_{\pm2.2}$ & 69.6$_{\pm2.4}$ & 68.8$_{\pm2.4}$ & 70.3$_{\pm2.3}$ \\
\midrule
\multicolumn{8}{l}{\textbf{Concept familiarity}} \\
Seen concepts & 71.2\% (2,064) & 78.2$_{\pm2.1}$ & 77.4$_{\pm2.2}$ & 79.2$_{\pm2.0}$ & 79.1$_{\pm2.1}$ & 79.3$_{\pm2.1}$ & 80.6$_{\pm2.0}$ \\
Unseen concepts & 28.8\% (834) & 16.7$_{\pm2.8}$ & 16.2$_{\pm2.4}$ & 18.8$_{\pm2.8}$ & 18.6$_{\pm2.7}$ & 19.3$_{\pm2.8}$ & 20.3$_{\pm2.8}$ \\
\midrule
\multicolumn{8}{l}{\textbf{Lexical match}} \\
Identical & 21.4\% (621) & 96.3$_{\pm1.4}$ & 96.1$_{\pm1.6}$ & 97.7$_{\pm1.4}$ & 96.8$_{\pm1.5}$ & 97.4$_{\pm1.2}$ & 96.8$_{\pm1.5}$ \\
Not identical & 78.6\% (2,277) & 50.8$_{\pm2.6}$ & 49.8$_{\pm2.8}$ & 52.0$_{\pm2.6}$ & 52.1$_{\pm2.9}$ & 52.3$_{\pm2.8}$ & 54.1$_{\pm2.8}$ \\
\midrule
\multicolumn{8}{l}{\textbf{Mention length}} \\
Single word & 27.6\% (800) & 93.4$_{\pm1.8}$ & 92.6$_{\pm2.4}$ & 93.5$_{\pm2.1}$ & 93.8$_{\pm2.0}$ & 93.9$_{\pm1.9}$ & 93.8$_{\pm2.1}$ \\
Multi-word & 72.4\% (2,098) & 48.0$_{\pm2.8}$ & 47.2$_{\pm2.9}$ & 49.7$_{\pm2.8}$ & 49.5$_{\pm2.8}$ & 49.9$_{\pm2.9}$ & 51.6$_{\pm2.9}$ \\
% $>$ 3 words & 41.3\% (1,197) & 35.3$_{\pm3.0}$ & 34.5$_{\pm2.9}$ & 36.9$_{\pm2.9}$ & 36.7$_{\pm3.2}$ & 37.4$_{\pm3.3}$ & 39.0$_{\pm3.4}$ \\
\midrule
\multicolumn{8}{l}{\textbf{Document recurrence}} \\
First occurrence & 87.8\% (2,545) & 58.0$_{\pm2.5}$ & 57.6$_{\pm2.6}$ & 59.4$_{\pm2.5}$ & 59.6$_{\pm2.7}$ & 59.8$_{\pm2.6}$ & 60.9$_{\pm2.6}$ \\
Recurring occurrence & 12.2\% (353) & 79.0$_{\pm5.0}$ & 75.6$_{\pm5.3}$ & 79.3$_{\pm4.8}$ & 76.5$_{\pm5.2}$ & 77.9$_{\pm5.1}$ & 80.2$_{\pm5.1}$ \\
2nd occurrence & 8.6\% (249) & 74.7$_{\pm5.7}$ & 70.3$_{\pm5.6}$ & 74.3$_{\pm5.7}$ & 71.5$_{\pm5.6}$ & 71.9$_{\pm6.0}$ & 74.7$_{\pm5.9}$ \\
3rd occurrence & 2.1\% (62) & 87.1$_{\pm8.7}$ & 85.5$_{\pm8.9}$ & 88.7$_{\pm8.3}$ & 83.9$_{\pm10.4}$ & 90.3$_{\pm7.9}$ & 90.3$_{\pm7.9}$ \\
4th occurrence & 0.7\% (21) & 90.5$_{\pm11.9}$ & 90.5$_{\pm11.9}$ & 90.5$_{\pm11.9}$ & 90.5$_{\pm11.9}$ & 90.5$_{\pm11.9}$ & 95.2$_{\pm8.4}$ \\
5th+ occurrence & 0.7\% (21) & 95.2$_{\pm8.8}$ & 95.2$_{\pm8.4}$ & 100.0$_{\pm0.0}$ & 100.0$_{\pm0.0}$ & 100.0$_{\pm0.0}$ & 100.0$_{\pm0.0}$ \\
\midrule
\multicolumn{8}{l}{\textbf{Concept structure}} \\
Simple concepts & 94.2\% (2,730) & 62.6$_{\pm2.5}$ & 61.8$_{\pm2.7}$ & 64.1$_{\pm2.4}$ & 63.7$_{\pm2.6}$ & 64.5$_{\pm2.5}$ & 65.1$_{\pm2.7}$ \\
Composite concepts & 5.8\% (168) & 26.8$_{\pm7.4}$ & 26.8$_{\pm7.1}$ & 24.4$_{\pm6.7}$ & 29.8$_{\pm7.0}$ & 21.4$_{\pm5.9}$ & 32.7$_{\pm7.1}$ \\
\bottomrule
\end{tabularx}

\caption{Detailed subset results on SympTEMIST. The first two rows report overall Recall@1 and Recall@5. All other rows report Recall@1 on specific mention subsets. The support column gives the percentage and number of examples in each subset. Each score is shown with half of the 95\% confidence interval. First occurrence refers to mentions whose gold concept has not appeared earlier in the document, while recurring occurrence refers to mentions whose gold concept has already appeared earlier. Simple and composite concept rows are reported only for the Spanish SPACCC datasets.}
\label{tab:detailed_symptemist}
\end{table*}

\begin{table*}
\centering
\small
\setlength{\tabcolsep}{1.4pt}
\renewcommand{\arraystretch}{1.1}
\begin{tabularx}{\textwidth}{>{\raggedright\arraybackslash}p{0.18\textwidth}rcccccc}
\toprule
\textbf{Subset} & \textbf{Support} & \textbf{Baseline-1B} & \textbf{LongBEL-1B} & \textbf{Ensemble-1B} & \textbf{Baseline-8B} & \textbf{LongBEL-8B} & \textbf{Ensemble-8B} \\
\midrule
\multicolumn{8}{l}{\textbf{Overall performance}} \\
Recall@1 overall & 100\% (2,341) & 62.5$_{\pm2.3}$ & 61.9$_{\pm2.4}$ & 64.3$_{\pm2.2}$ & 63.2$_{\pm2.5}$ & 63.6$_{\pm2.1}$ & 65.8$_{\pm2.2}$ \\
Recall@5 overall & 100\% (2,341) & 69.5$_{\pm2.1}$ & 68.9$_{\pm2.1}$ & 73.1$_{\pm2.1}$ & 71.1$_{\pm2.2}$ & 69.9$_{\pm2.1}$ & 73.2$_{\pm2.1}$ \\
\midrule
\multicolumn{8}{l}{\textbf{Concept familiarity}} \\
Seen concepts & 61.7\% (1,445) & 83.0$_{\pm2.2}$ & 80.7$_{\pm2.4}$ & 83.9$_{\pm2.1}$ & 83.8$_{\pm2.2}$ & 84.6$_{\pm1.9}$ & 86.4$_{\pm1.8}$ \\
Unseen concepts & 38.3\% (896) & 29.2$_{\pm3.2}$ & 31.5$_{\pm3.3}$ & 32.8$_{\pm3.4}$ & 30.0$_{\pm3.5}$ & 29.9$_{\pm3.4}$ & 32.5$_{\pm3.7}$ \\
\midrule
\multicolumn{8}{l}{\textbf{Lexical match}} \\
Identical & 24.9\% (583) & 94.2$_{\pm1.9}$ & 93.1$_{\pm2.0}$ & 95.2$_{\pm1.9}$ & 93.3$_{\pm2.2}$ & 93.5$_{\pm2.0}$ & 95.0$_{\pm1.8}$ \\
Not identical & 75.1\% (1,758) & 51.9$_{\pm2.6}$ & 51.5$_{\pm2.7}$ & 54.1$_{\pm2.5}$ & 53.2$_{\pm2.8}$ & 53.8$_{\pm2.5}$ & 56.1$_{\pm2.6}$ \\
\midrule
\multicolumn{8}{l}{\textbf{Mention length}} \\
Single word & 28.3\% (663) & 86.7$_{\pm3.4}$ & 85.2$_{\pm3.5}$ & 87.6$_{\pm3.4}$ & 86.0$_{\pm3.5}$ & 87.6$_{\pm3.1}$ & 88.2$_{\pm3.1}$ \\
Multi-word & 71.7\% (1,678) & 52.9$_{\pm2.7}$ & 52.6$_{\pm2.8}$ & 55.1$_{\pm2.6}$ & 54.2$_{\pm2.9}$ & 54.2$_{\pm2.6}$ & 56.9$_{\pm2.8}$ \\
% $>$ 3 words & 29.9\% (701) & 31.8$_{\pm3.5}$ & 33.0$_{\pm3.6}$ & 35.4$_{\pm3.6}$ & 31.8$_{\pm3.5}$ & 33.2$_{\pm3.4}$ & 35.2$_{\pm3.5}$ \\
\midrule
\multicolumn{8}{l}{\textbf{Document recurrence}} \\
First occurrence & 85.0\% (1,991) & 61.6$_{\pm2.4}$ & 60.7$_{\pm2.5}$ & 62.9$_{\pm2.3}$ & 62.2$_{\pm2.7}$ & 61.8$_{\pm2.1}$ & 63.9$_{\pm2.4}$ \\
Recurring occurrence & 15.0\% (350) & 67.4$_{\pm5.3}$ & 68.6$_{\pm5.6}$ & 72.3$_{\pm5.1}$ & 68.9$_{\pm5.3}$ & 74.3$_{\pm4.8}$ & 76.3$_{\pm4.6}$ \\
2nd occurrence & 11.5\% (269) & 66.5$_{\pm5.6}$ & 67.3$_{\pm5.6}$ & 71.0$_{\pm5.0}$ & 68.8$_{\pm5.3}$ & 72.9$_{\pm5.2}$ & 75.8$_{\pm5.1}$ \\
3rd occurrence & 2.4\% (56) & 67.9$_{\pm10.8}$ & 71.4$_{\pm10.5}$ & 76.8$_{\pm9.9}$ & 67.9$_{\pm10.8}$ & 75.0$_{\pm9.6}$ & 75.0$_{\pm9.6}$ \\
4th occurrence & 0.6\% (13) & 69.2$_{\pm23.9}$ & 69.2$_{\pm26.7}$ & 69.2$_{\pm23.9}$ & 76.9$_{\pm25.0}$ & 76.9$_{\pm25.0}$ & 76.9$_{\pm25.0}$ \\
5th+ occurrence & 0.5\% (12) & 83.3$_{\pm17.4}$ & 83.3$_{\pm17.4}$ & 83.3$_{\pm17.4}$ & 66.7$_{\pm40.0}$ & 100.0$_{\pm0.0}$ & 91.7$_{\pm8.7}$ \\
\midrule
\multicolumn{8}{l}{\textbf{Concept structure}} \\
Simple concepts & 96.4\% (2,257) & 64.3$_{\pm2.2}$ & 63.7$_{\pm2.4}$ & 66.3$_{\pm2.2}$ & 65.0$_{\pm2.4}$ & 65.6$_{\pm2.0}$ & 67.7$_{\pm2.1}$ \\
Composite concepts & 3.6\% (84) & 13.1$_{\pm7.5}$ & 13.1$_{\pm7.1}$ & 11.9$_{\pm6.8}$ & 16.7$_{\pm8.4}$ & 11.9$_{\pm6.5}$ & 15.5$_{\pm7.5}$ \\
\bottomrule
\end{tabularx}

\caption{Detailed subset results on DisTEMIST. The first two rows report overall Recall@1 and Recall@5. All other rows report Recall@1 on specific mention subsets. The support column gives the percentage and number of examples in each subset. Each score is shown with half of the 95\% confidence interval. First occurrence refers to mentions whose gold concept has not appeared earlier in the document, while recurring occurrence refers to mentions whose gold concept has already appeared earlier. Simple and composite concept rows are reported only for the Spanish SPACCC datasets.}
\label{tab:detailed_distemist}
\end{table*}

\begin{table*}[t]
\centering
\small
\setlength{\tabcolsep}{1.4pt}
\renewcommand{\arraystretch}{1.1}
\begin{tabularx}{\textwidth}{>{\raggedright\arraybackslash}p{0.18\textwidth}rcccccc}
\toprule
\textbf{Subset} & \textbf{Support} & \textbf{Baseline-1B} & \textbf{LongBEL-1B} & \textbf{Ensemble-1B} & \textbf{Baseline-8B} & \textbf{LongBEL-8B} & \textbf{Ensemble-8B} \\
\midrule
\multicolumn{8}{l}{\textbf{Overall performance}} \\
Recall@1 overall & 100\% (3,582) & 67.4$_{\pm2.1}$ & 66.6$_{\pm2.1}$ & 69.0$_{\pm2.0}$ & 68.3$_{\pm2.2}$ & 69.0$_{\pm2.1}$ & 71.0$_{\pm2.0}$ \\
Recall@5 overall & 100\% (3,582) & 72.9$_{\pm1.9}$ & 71.7$_{\pm2.1}$ & 76.1$_{\pm1.9}$ & 74.8$_{\pm1.8}$ & 74.4$_{\pm1.9}$ & 75.7$_{\pm2.0}$ \\
\midrule
\multicolumn{8}{l}{\textbf{Concept familiarity}} \\
Seen concepts & 73.6\% (2,636) & 84.3$_{\pm1.8}$ & 83.5$_{\pm1.8}$ & 85.9$_{\pm1.7}$ & 86.2$_{\pm1.6}$ & 86.4$_{\pm1.6}$ & 87.6$_{\pm1.6}$ \\
Unseen concepts & 26.4\% (946) & 20.1$_{\pm2.8}$ & 19.7$_{\pm2.8}$ & 21.8$_{\pm3.0}$ & 18.6$_{\pm2.9}$ & 20.6$_{\pm2.9}$ & 24.9$_{\pm3.2}$ \\
\midrule
\multicolumn{8}{l}{\textbf{Lexical match}} \\
Identical & 18.5\% (664) & 97.4$_{\pm1.4}$ & 94.9$_{\pm2.0}$ & 97.9$_{\pm1.6}$ & 97.0$_{\pm1.5}$ & 96.5$_{\pm1.8}$ & 98.8$_{\pm1.0}$ \\
Not identical & 81.5\% (2,918) & 60.5$_{\pm2.3}$ & 60.2$_{\pm2.4}$ & 62.4$_{\pm2.3}$ & 61.9$_{\pm2.4}$ & 62.6$_{\pm2.4}$ & 64.7$_{\pm2.3}$ \\
\midrule
\multicolumn{8}{l}{\textbf{Mention length}} \\
Single word & 33.6\% (1,202) & 84.7$_{\pm2.8}$ & 83.9$_{\pm2.9}$ & 86.3$_{\pm2.6}$ & 85.9$_{\pm2.5}$ & 86.8$_{\pm2.4}$ & 87.6$_{\pm2.5}$ \\
Multi-word & 66.4\% (2,380) & 58.6$_{\pm2.6}$ & 57.9$_{\pm2.7}$ & 60.2$_{\pm2.6}$ & 59.5$_{\pm2.6}$ & 60.1$_{\pm2.6}$ & 62.7$_{\pm2.7}$ \\
% $>$ 3 words & 26.2\% (940) & 33.6$_{\pm3.3}$ & 32.0$_{\pm3.4}$ & 36.4$_{\pm3.3}$ & 35.9$_{\pm3.4}$ & 35.7$_{\pm3.0}$ & 38.2$_{\pm3.1}$ \\
\midrule
\multicolumn{8}{l}{\textbf{Document recurrence}} \\
First occurrence & 84.4\% (3,023) & 66.2$_{\pm2.1}$ & 65.1$_{\pm2.2}$ & 67.6$_{\pm2.1}$ & 67.1$_{\pm2.1}$ & 67.4$_{\pm2.2}$ & 69.6$_{\pm2.1}$ \\
Recurring occurrence & 15.6\% (559) & 73.7$_{\pm4.5}$ & 74.8$_{\pm4.2}$ & 76.0$_{\pm4.1}$ & 75.3$_{\pm4.3}$ & 77.6$_{\pm3.8}$ & 78.7$_{\pm4.1}$ \\
2nd occurrence & 11.5\% (412) & 73.3$_{\pm4.2}$ & 73.5$_{\pm4.7}$ & 74.8$_{\pm4.4}$ & 74.0$_{\pm4.3}$ & 74.5$_{\pm4.0}$ & 76.7$_{\pm4.2}$ \\
3rd occurrence & 2.7\% (95) & 78.9$_{\pm8.0}$ & 77.9$_{\pm7.6}$ & 80.0$_{\pm7.4}$ & 77.9$_{\pm8.3}$ & 86.3$_{\pm6.1}$ & 84.2$_{\pm6.7}$ \\
4th occurrence & 0.9\% (34) & 67.6$_{\pm15.2}$ & 76.5$_{\pm14.6}$ & 79.4$_{\pm13.3}$ & 79.4$_{\pm14.7}$ & 85.3$_{\pm12.7}$ & 85.3$_{\pm12.7}$ \\
5th+ occurrence & 0.5\% (18) & 66.7$_{\pm21.9}$ & 83.3$_{\pm20.6}$ & 77.8$_{\pm22.2}$ & 83.3$_{\pm22.2}$ & 88.9$_{\pm17.6}$ & 83.3$_{\pm22.2}$ \\
\midrule
\multicolumn{8}{l}{\textbf{Concept structure}} \\
Simple concepts & 97.9\% (3,508) & 68.3$_{\pm2.2}$ & 67.5$_{\pm2.1}$ & 69.9$_{\pm2.1}$ & 69.4$_{\pm2.1}$ & 70.0$_{\pm2.1}$ & 72.0$_{\pm2.0}$ \\
Composite concepts & 2.1\% (74) & 21.6$_{\pm8.9}$ & 25.7$_{\pm9.3}$ & 24.3$_{\pm9.5}$ & 17.6$_{\pm7.1}$ & 21.6$_{\pm8.9}$ & 24.3$_{\pm9.3}$ \\
\bottomrule
\end{tabularx}

\caption{Detailed subset results on MedProcNER. The first two rows report overall Recall@1 and Recall@5. All other rows report Recall@1 on specific mention subsets. The support column gives the percentage and number of examples in each subset. Each score is shown with half of the 95\% confidence interval. First occurrence refers to mentions whose gold concept has not appeared earlier in the document, while recurring occurrence refers to mentions whose gold concept has already appeared earlier. Simple and composite concept rows are reported only for the Spanish SPACCC datasets.}
\label{tab:detailed_medprocner}
\end{table*}

\section{Qualitative Examples}
\label{sec:appendix_qual_examples}

Table~\ref{tab:qualitative_examples} compares LongBEL-8B with the sentence-level Llama-8B baseline on representative cases: cases where document-level context helps, cases where memory propagates an error, and cases where both systems either fail or succeed.

\begin{table*}
\centering
\small
\setlength{\tabcolsep}{6pt}
\renewcommand{\arraystretch}{1}

\begin{tabularx}{\textwidth}{
>{\raggedright\arraybackslash}p{0.11\textwidth}
>{\raggedright\arraybackslash}X}
\toprule
\textbf{Outcome} & \textbf{Example} \\
\midrule

\textbf{LongBEL} \cmarkcol
\newline
\newline
\textbf{Baseline} \xmarkcol
&
\begin{minipage}[t]{\linewidth}
\textbf{Mention:} [ND]

\textbf{Local context:} Therefore, we investigated factors associated with [ND] in patients with LAA.

\textbf{Relevant global context:} In some patients with acute ischemic stroke, neurological deterioration (ND) is observed, and it is difficult to predict at the time of admission.

\textbf{Previous normalized mentions:}
\begin{list}{}{\leftmargin=3em \itemsep=0pt \parsep=0pt \topsep=1pt \partopsep=0pt}
    \item neurological deterioration $\rightarrow$ Neurological status deteriorated
    \item ND $\rightarrow$ Neurological status deteriorated
    \item large-artery atherosclerosis $\rightarrow$ Large artery atherosclerosis
\end{list}

\textbf{Gold:} Neurological status deteriorated (C1536136)

\textbf{Baseline:} Nicotine dependence (disorder) (C0028043) \xmarkcol

\textbf{LongBEL-8B:} Neurological status deteriorated (C1536136) \cmarkcol

\textit{Takeaway:} The abbreviation ``ND'' is explicitly defined earlier in the document as neurological deterioration. The sentence-level baseline only sees the ambiguous abbreviation and links it to nicotine dependence, whereas LongBEL uses the previous abbreviation definition to recover the correct concept.
\end{minipage}
\\
\midrule

\textbf{LongBEL} \cmarkcol
\newline
\newline
\textbf{Baseline} \xmarkcol
&
\begin{minipage}[t]{\linewidth}
\textbf{Mention:} [PET]

\textbf{Local context:} She was admitted overnight with suspected [PET] and was started on urgent treatment.

\textbf{Relevant global context:} A 29-year-old pregnant woman presented with severe-range hypertension, headache, and epigastric pain. Laboratory testing showed proteinuria and mildly elevated liver enzymes.

\textbf{Previous normalized mentions:}
\begin{list}{}{\leftmargin=3em \itemsep=0pt \parsep=0pt \topsep=1pt \partopsep=0pt}
    \item pregnant woman $\rightarrow$ Pregnant Woman
    \item severe-range hypertension $\rightarrow$ Hypertensive disease
    \item proteinuria $\rightarrow$ Proteinurias
\end{list}

\textbf{Gold:} Pre-eclamptic toxemia (C0032914)

\textbf{Baseline:} Petechial haemorrhage (C0031256) \xmarkcol

\textbf{LongBEL-8B:} Pre-eclamptic toxemia (C0032914) \cmarkcol

\textit{Takeaway:} Unlike an explicit abbreviation-definition case, this example requires combining several document-level clinical cues. The local sentence only contains the ambiguous abbreviation ``PET'', while LongBEL uses the obstetric context and previous normalized mentions, especially ``pregnant Woman'', ``hypertensive disease'', and ``proteinurias'', to recover the correct concept.
\end{minipage}
\\
\midrule

\textbf{LongBEL} \xmarkcol
\newline
\newline
\textbf{Baseline} \cmarkcol
&
\begin{minipage}[t]{\linewidth}
\textbf{Mention:} [IFN-$\gamma$]

\textbf{Local context:} The aim of the study is to investigate the association of [IFN-$\gamma$] and IL-10 gene single nucleotide polymorphisms.

\textbf{Global context:} (...) Interferon gamma (IFN-$\gamma$) and interleukin 10 (IL-10) polymorphisms in Chinese children (...)

\textbf{Previous normalized mentions:}
\begin{list}{}{\leftmargin=3em \itemsep=0pt \parsep=0pt \topsep=1pt \partopsep=0pt}
    \item Interferon gamma $\rightarrow$ Interleukin 9 Gene
    \item IFN-$\gamma$ $\rightarrow$ Interleukin 9 Gene
    \item interleukin 10 $\rightarrow$ Interleukin 10
    \item IL-10 $\rightarrow$ Interleukin 10
\end{list}

\textbf{Gold:} IFNG gene (C1334085)

\textbf{Baseline:} IFNG gene (C1334085) \cmarkcol

\textbf{LongBEL-8B:} Interleukin 9 Gene (C1334127) \xmarkcol

\textit{Takeaway:} Although the document contains an explicit definition, LongBEL mislinks the first occurrence  ``Interferon gamma (IFN-$\gamma$)'' and then propagates this incorrect prediction through memory to the later mention.
\end{minipage}
\\
\midrule

\textbf{LongBEL} \xmarkcol
\newline
\newline
\textbf{Baseline} \xmarkcol
&
\begin{minipage}[t]{\linewidth}
\textbf{Mention:} [MK-8(H2)]

\textbf{Local context:} The major menaquinone were [MK-8(H2)] (72\%) and MK-9(H2) (28\%).

\textbf{Gold:} vitamin MK 8 (C0084980)

\textbf{Baseline:} MAPKAPK2 protein, human (C1452441) \xmarkcol

\textbf{LongBEL-8B:} vitamin K2 (C0086605) \xmarkcol

\textit{Takeaway:} The mention is a specialized chemical abbreviation with no additional disambiguating evidence elsewhere in the document. LongBEL predicts a related vitamin K concept, but both models fail to recover the exact KB concept.
\end{minipage}
\\
\midrule

\textbf{LongBEL} \cmarkcol
\newline
\newline
\textbf{Baseline} \cmarkcol
&
\begin{minipage}[t]{\linewidth}
\textbf{Mention:} [DDIs]

\textbf{Local context:} Elderly patients are at high risk from drug-drug interactions ([DDIs]).

\textbf{Gold:} Drug Interactions (C0687133)

\textbf{Baseline:} Drug Interactions (C0687133) \cmarkcol

\textbf{LongBEL-8B:} Drug Interactions (C0687133) \cmarkcol

\textit{Takeaway:} The local sentence explicitly defines the abbreviation, so both the sentence-level baseline and LongBEL resolve the mention correctly.
\end{minipage}
\\
\bottomrule
\end{tabularx}

\caption{
Qualitative examples comparing the sentence-level Llama-8B baseline and LongBEL-8B. The examples show when document-level context helps, when memory propagates an error, and when both systems fail or succeed. All examples are from MedMentions-ST21pv except the manually constructed PET example.
}
\label{tab:qualitative_examples}
\end{table*}

\section{Cross-Dataset Generalization with Source-KB Decoding}
\label{sec:cross_dataset_kb}

The main experiment uses the KB associated with the evaluation dataset. Here, we consider a stricter setting in which both the trained model and its source-KB decoding trie are transferred unchanged. We also report the proportion of target concepts covered by the source KB.

\begin{table}
\centering
\small
\setlength{\tabcolsep}{4pt}
\renewcommand{\arraystretch}{1.08}
\resizebox{\columnwidth}{!}{%
\begin{tabular}{llccc}
\toprule
\shortstack{\textbf{Training}\\\textbf{dataset}} &
\shortstack{\textbf{System /}\\\textbf{metric}} &
\multicolumn{3}{c}{\textbf{Evaluation dataset}} \\
\cmidrule(lr){3-5}
& &
\shortstack{\textbf{MM-}\\\textbf{ST21pv}} &
\shortstack{\textbf{QUAERO-}\\\textbf{EMEA}} &
\textbf{SPACCC} \\
\midrule
\multirow{3}{*}{\shortstack{\textbf{MM-}\\\textbf{ST21pv}}}
& Baseline
& $75.0 {\scriptstyle \pm 0.9}$
& $55.9 {\scriptstyle \pm 4.9}$
& $35.5 {\scriptstyle \pm 1.3}$ \\
& LongBEL
& $\mathbf{79.3} {\scriptstyle \pm 0.8}$
& $\mathbf{63.6} {\scriptstyle \pm 4.0}$
& $\mathbf{39.6} {\scriptstyle \pm 1.4}$ \\
& \textit{KB coverage}
& $100.0\%$
& $96.8\%$
& $95.7\%$ \\
\midrule
\multirow{3}{*}{\shortstack{\textbf{QUAERO-}\\\textbf{EMEA}}}
& Baseline
& $48.1 {\scriptstyle \pm 1.0}$
& $73.8 {\scriptstyle \pm 4.0}$
& $31.5 {\scriptstyle \pm 1.3}$ \\
& LongBEL
& $\mathbf{53.3} {\scriptstyle \pm 1.1}$
& $\mathbf{75.4} {\scriptstyle \pm 3.4}$
& $\mathbf{33.5} {\scriptstyle \pm 1.3}$ \\
& \textit{KB coverage}
& $96.9\%$
& $100.0\%$
& $94.0\%$ \\
\midrule
\multirow{3}{*}{\textbf{SPACCC}}
& Baseline
& $8.1 {\scriptstyle \pm 0.6}$
& $10.5 {\scriptstyle \pm 2.0}$
& $65.1 {\scriptstyle \pm 1.5}$ \\
& LongBEL
& $\mathbf{9.2} {\scriptstyle \pm 0.7}$
& $\mathbf{10.7} {\scriptstyle \pm 2.0}$
& $\mathbf{65.3} {\scriptstyle \pm 1.5}$ \\
& \textit{KB coverage}
& $34.5\%$
& $48.0\%$
& $100.0\%$ \\
\bottomrule
\end{tabular}%
}
\caption{Cross-dataset Recall@1 with source-KB decoding. Models and decoding tries are transferred unchanged from the row dataset to the column dataset. KB coverage is the proportion of distinct evaluation concepts present in the source KB.}
\label{tab:cross_dataset_source_kb}
\end{table}

LongBEL has a higher point estimate than the Baseline in all six transfer directions, with gains from 0.2 to 7.7 points. Transfer from SPACCC is particularly weak because its restricted KB covers only 34.5\% of MM-ST21pv concepts and 48.0\% of QUAERO-EMEA concepts. Conversely, the UMLS-based KBs cover at least 94\% of SPACCC concepts, but transfer remains below within-dataset performance. KB coverage therefore explains only part of the transfer gap; differences in language, corpus, and annotation guidelines also play an important role.

\section{Robustness to Noisy Memory}
\label{sec:appendix_copy_error}

To evaluate whether a model propagates errors through memory, we define the \textit{Copy Wrong Memory Error Rate} (CWME). CWME measures how often a model repeats an earlier incorrect prediction for the same ground-truth concept.

Let $M_e = (m_1, m_2, \dots, m_k)$ be the ordered mentions of concept $e$ in a document. Suppose the model first predicts an incorrect concept $e' \neq e$ at mention $m_f$. All later mentions of $e$ are then \textit{exposed mentions}, since the wrong prediction is present in memory:
\begin{equation*}
    E_e = \{ m_i \in M_e \mid i > f \}.
\end{equation*}

An \textit{echoed error} occurs when the model repeats the same wrong prediction $e'$:
\begin{equation*}
    R_e = \{ m_i \in E_e \mid \hat{e}_i = e' \}.
\end{equation*}

\begin{figure}[t]
    \centering
    \includegraphics[width=0.5\textwidth]{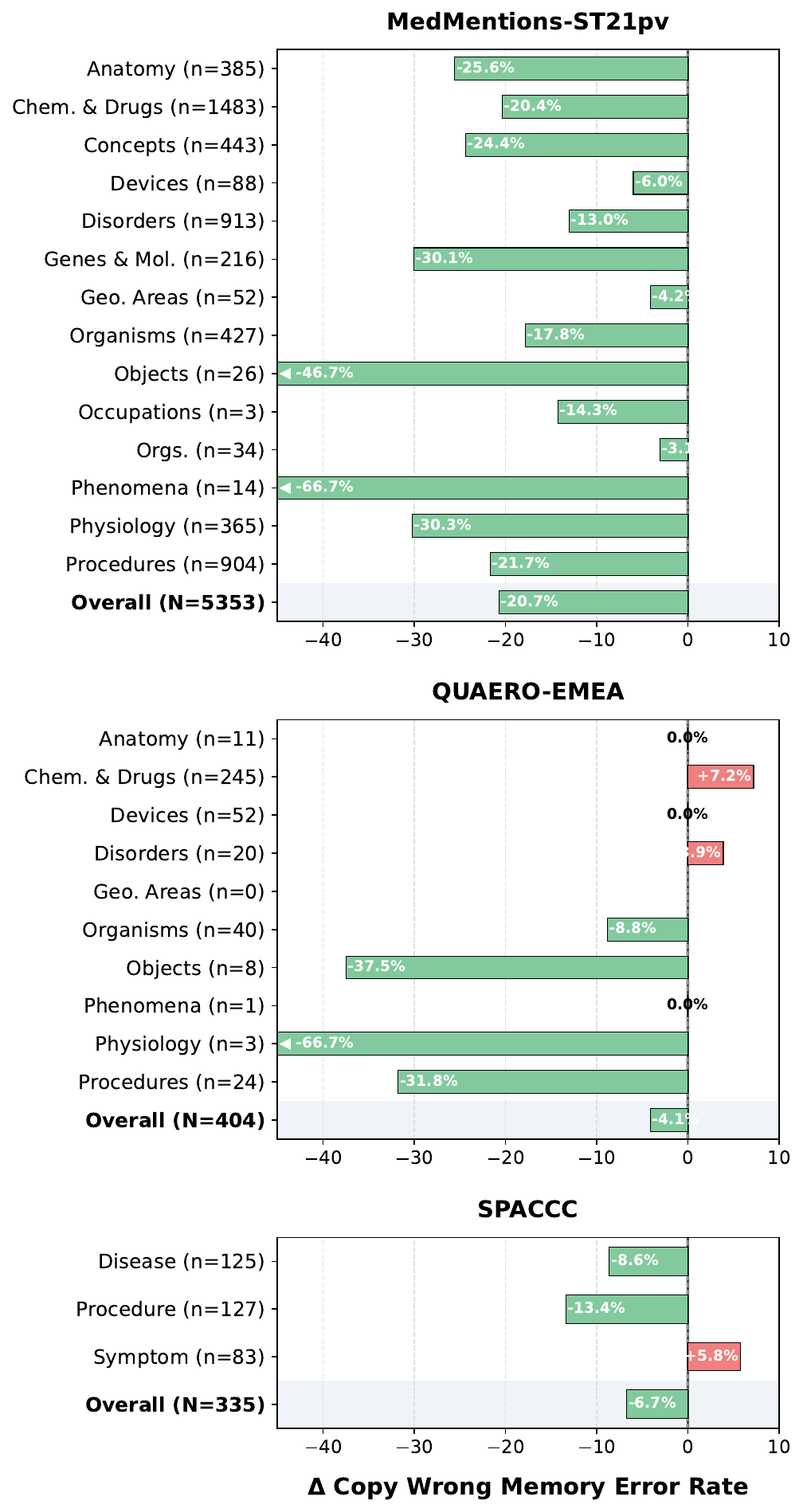}
    \caption{
    Change in Copy Wrong Memory Error Rate when using robust memory instead of gold memory for LongBEL-1B. Negative values indicate fewer cascading errors. Results are computed on exposed recurring concepts and reported by semantic group.
    }
    \label{fig:copy_error_1b}
\end{figure}

\begin{figure}[t]
    \centering
    \includegraphics[width=0.5\textwidth]{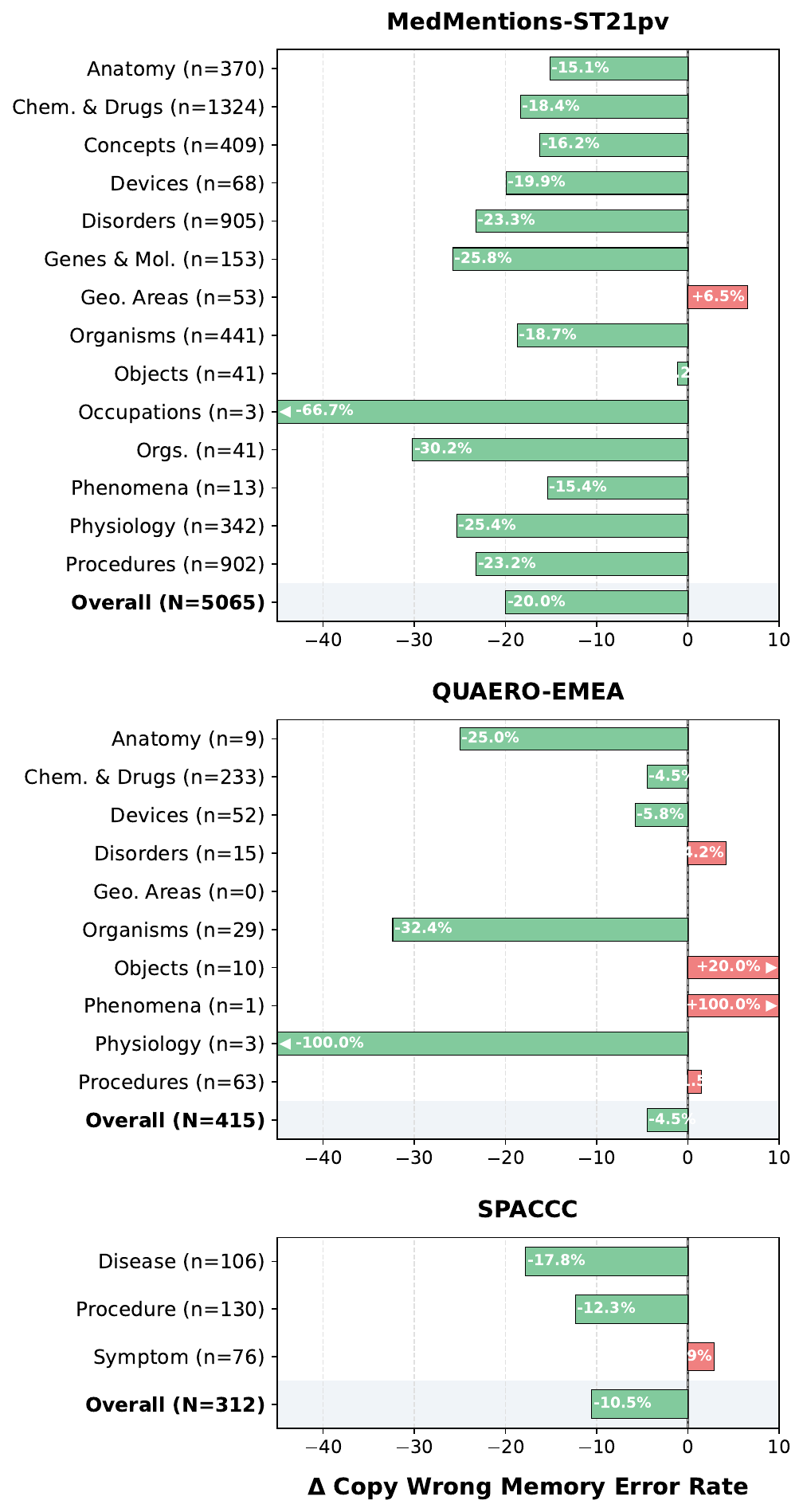}
    \caption{
    Change in Copy Wrong Memory Error Rate when using robust memory instead of gold memory for LongBEL-8B. Negative values indicate fewer cascading errors. Results are computed on exposed recurring concepts and reported by semantic group.
    }
    \label{fig:copy_error_8b}
\end{figure}

CWME is computed as:
\begin{equation*}
    \text{CWME} =
    \frac{\sum_{e \in \mathcal{E}} |R_e|}
         {\sum_{e \in \mathcal{E}} |E_e|}
    \times 100,
\end{equation*}
where $\mathcal{E}$ includes concepts with an initial error and at least one later exposed mention. Lower CWME indicates fewer cascading errors. Figures~\ref{fig:copy_error_1b} and~\ref{fig:copy_error_8b} compare gold-memory training with robust cross-validated memory training using
\begin{equation*}
    \Delta \text{CWME}
    =
    \text{CWME}_{\text{robust memory}}
    -
    \text{CWME}_{\text{gold memory}}.
\end{equation*}
Negative values indicate fewer copied errors with robust memory.

Table~\ref{tab:memory_training} complements CWME with an overall Recall@1 comparison for LongBEL-8B. Both settings use predicted memory at inference, so only the training memory differs. Robust memory slightly improves Recall@1 on all three datasets.

\begin{table}[t]
\centering
\small
\begin{tabular}{lcc}
\toprule
\textbf{Dataset}
& \shortstack{\textbf{Gold}\\\textbf{memory}}
& \shortstack{\textbf{Robust}\\\textbf{memory}} \\
\midrule
\multicolumn{3}{l}{\textit{LongBEL-1B}} \\
MM-ST21pv & $77.4 {\scriptstyle \pm 0.9}$ & $\mathbf{77.6} {\scriptstyle \pm 0.9}$ \\
QUAERO-EMEA & $\mathbf{75.2} {\scriptstyle \pm 4.3}$ & $74.5 {\scriptstyle \pm 3.7}$ \\
SPACCC & $62.8 {\scriptstyle \pm 1.6}$ & $\mathbf{63.1} {\scriptstyle \pm 1.6}$ \\
\midrule
\multicolumn{3}{l}{\textit{LongBEL-8B}} \\
MM-ST21pv & $78.9 {\scriptstyle \pm 0.9}$ & $\mathbf{79.3} {\scriptstyle \pm 0.8}$ \\
QUAERO-EMEA & $75.2 {\scriptstyle \pm 3.1}$ & $\mathbf{75.4} {\scriptstyle \pm 3.4}$ \\
SPACCC & $64.5 {\scriptstyle \pm 1.7}$ & $\mathbf{65.3} {\scriptstyle \pm 1.5}$ \\
\bottomrule
\end{tabular}
\caption{Recall@1 with gold or robust training memory. Both settings use predicted memory at inference. Values are percentages with half of the 95\% confidence interval.}
\label{tab:memory_training}
\end{table}

Overall, robust memory consistently reduces copied errors. For LongBEL-1B, CWME decreases by 20.7 points on MM-ST21pv, 4.1 points on QUAERO-EMEA, and 6.7 points on SPACCC. For LongBEL-8B, the corresponding reductions are 20.0, 4.5, and 10.5 points, respectively. These results show that training with predicted memory makes the model less likely to repeat earlier mistakes. The reduction is strongest and most stable on MM-ST21pv, where many semantic groups contain enough exposed mentions to provide reliable estimates. In contrast, some groups in QUAERO-EMEA and SPACCC contain very few examples, leading to noisier category-level results and occasional positive deltas. This is consistent with our main analysis: robust memory is most effective when enough recurring concepts are available to learn reliable consistency patterns.

\section{Saliency-based Analysis}
\label{sec:appendix_saliency}

\begin{figure*}[t]
    \centering
    \footnotesize

    \begin{minipage}{0.93\textwidth}
        \centering
        \textbf{(a) ND: using an earlier abbreviation definition} \\[2pt]
        \includegraphics[width=\textwidth]{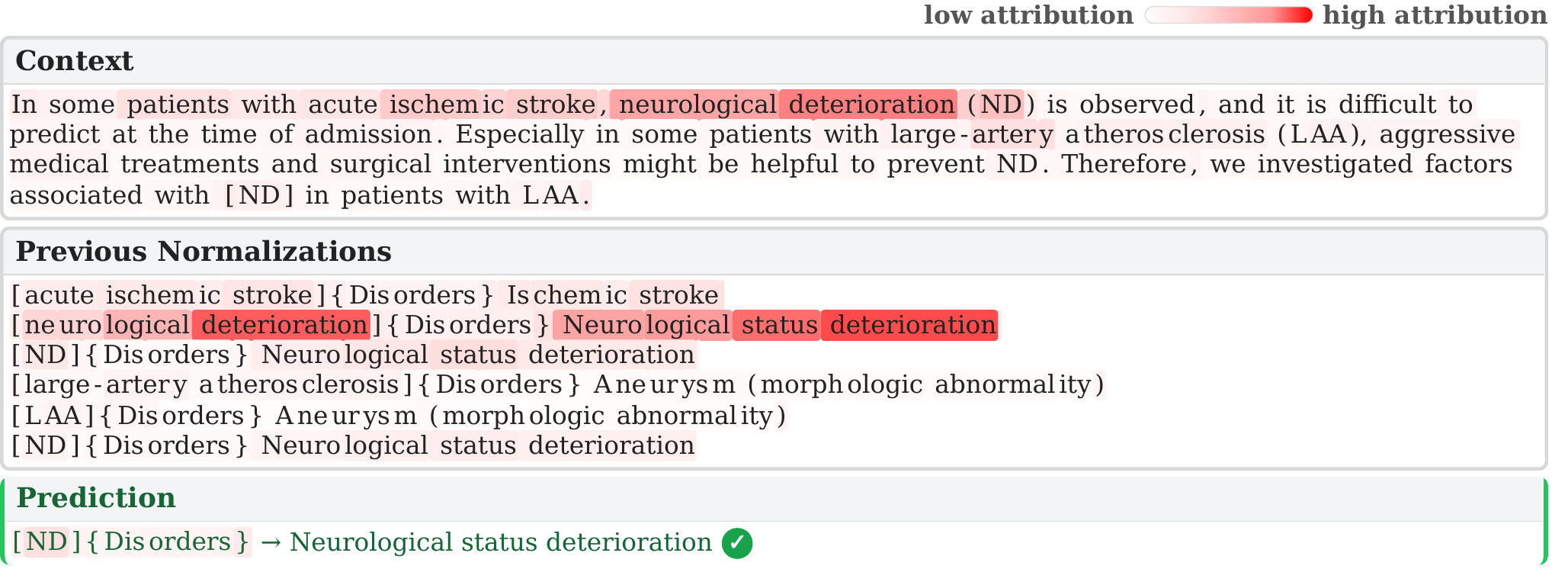}
        \vspace{-2pt}

        \raggedright
        LongBEL-8B gives high saliency to the earlier definition ``neurological deterioration (ND)'' and to the corresponding memory entries. This helps the model normalize the later mention of ND correctly.
    \end{minipage}

    \vspace{0.35em}

    \begin{minipage}{0.93\textwidth}
        \centering
        \textbf{(b) PET: using clinical context and memory} \\[2pt]
        \includegraphics[width=\textwidth]{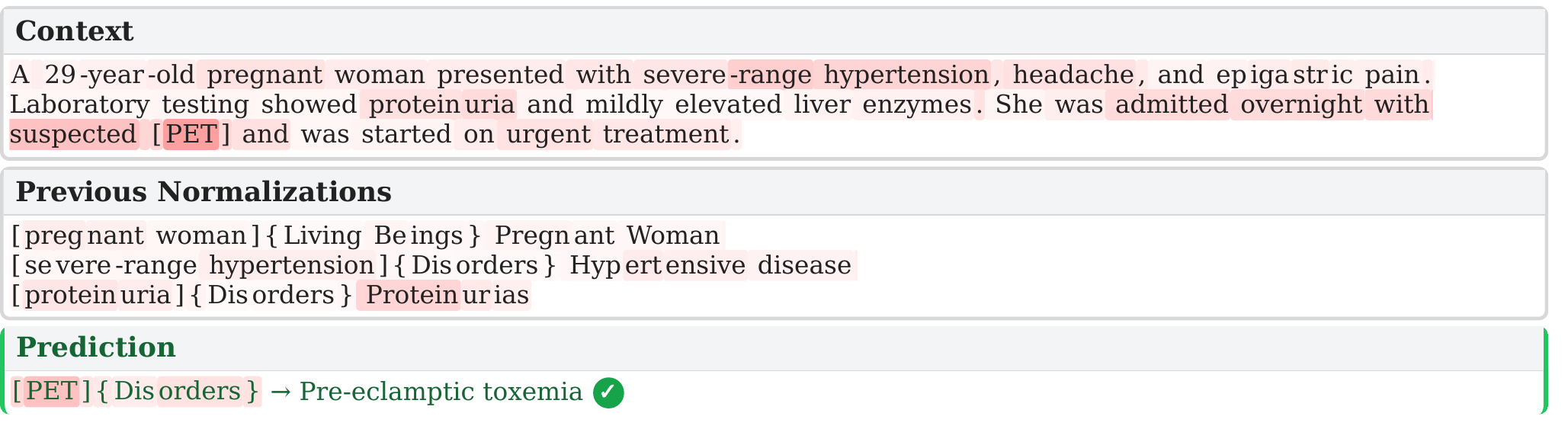}
        \vspace{-2pt}

        \raggedright
        LongBEL-8B uses the obstetric context and previous normalized mentions, especially hypertension and proteinuria, to link PET to pre-eclamptic toxemia.
    \end{minipage}

    \vspace{0.35em}

    \begin{minipage}{0.93\textwidth}
        \centering
        \textbf{(c) IFN-$\gamma$: when memory propagates an error} \\[2pt]
        \includegraphics[width=\textwidth]{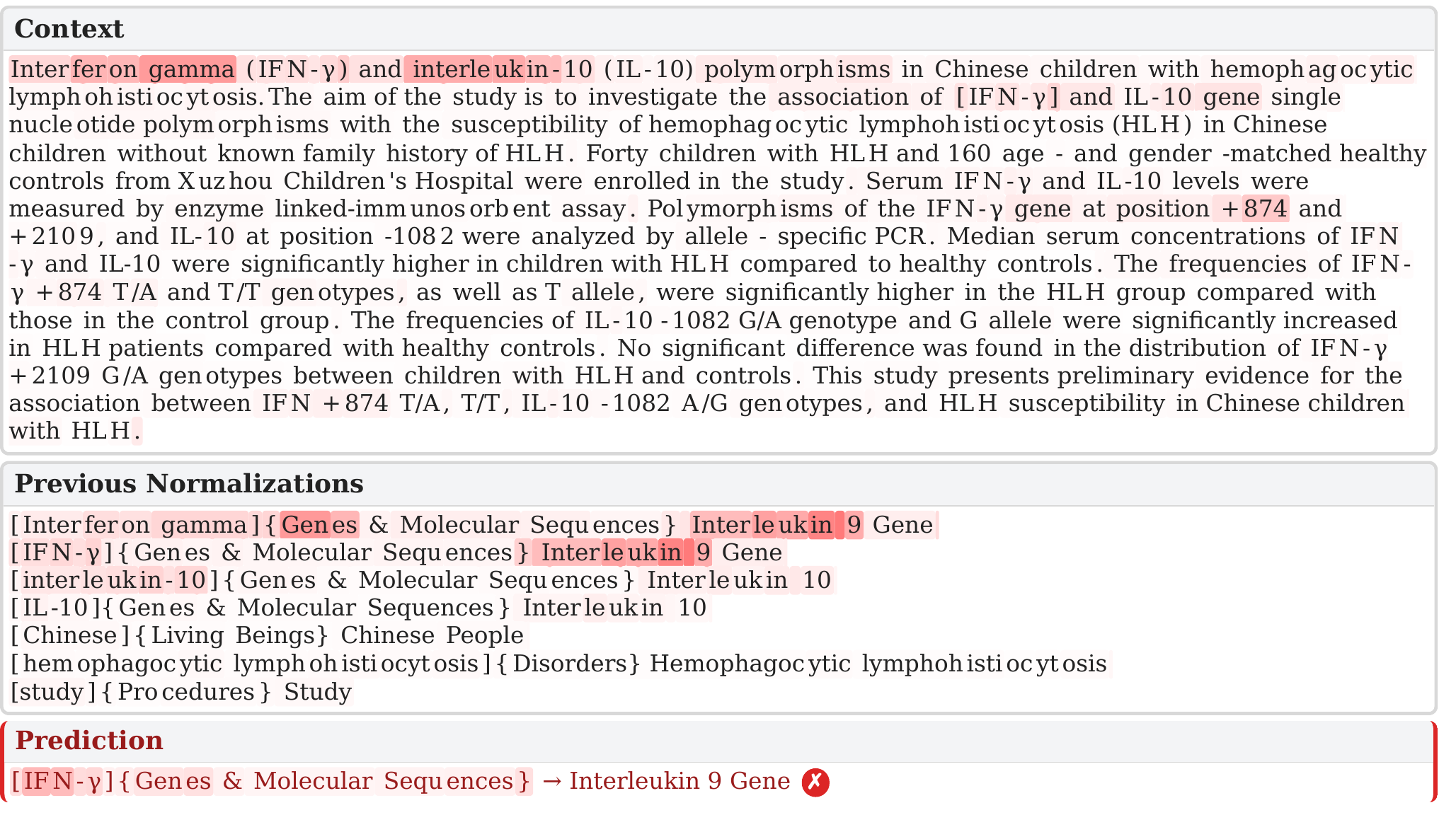}
        \vspace{-2pt}

        \raggedright
        LongBEL-8B gives high saliency to an incorrect previous normalization. This shows that an early linking error can influence a later abbreviation prediction through memory.
    \end{minipage}

    \caption{
    Saliency maps for three examples. Panels~(a) and~(c) are from MedMentions-ST21pv. Panel~(b) is a manually constructed example. Panel~(a) shows how an earlier abbreviation definition helps LongBEL-8B make the correct prediction. Panel~(b) shows how clinical context and previous normalized mentions help the model resolve an ambiguous abbreviation. Panel~(c) shows a failure where memory propagates an incorrect earlier prediction.
    }
    \label{fig:saliency_maps}
\end{figure*}

Figure~\ref{fig:saliency_maps} shows token-level saliency maps for three examples, with all predictions and saliency scores produced by LongBEL-8B. The maps show that the model can use information from document context and previous predictions, which can help but can also propagate earlier errors.

We compute saliency with a simple gradient-based method. After decoding, we keep the top generated sequence. We define the objective as the sum of the log-probabilities of the generated output tokens:
\[
S(x, y) = \sum_{t \in y} \log p_\theta(y_t \mid x, y_{<t}),
\]
where $x$ is the prompt and $y$ is the generated output. We then compute the gradient of this objective with respect to the input token embeddings. The saliency score of an input token is the $\ell_2$ norm of this gradient:
\[
a_i = \left\| \frac{\partial S(x, y)}{\partial e_i} \right\|_2,
\]
where $e_i$ is the embedding of the $i$-th input token. We only show scores for prompt tokens. Generated tokens and padding tokens are not shown. For each example, we normalize the scores by the maximum score in that example.

This saliency analysis is a local sensitivity measure. A highlighted token is a token that strongly affects the likelihood of the generated prediction. It is not based on attention weights, and it should not be read as a causal explanation.

\section{Inference Parallelization}
\label{sec:inference_parallelization}

Although LongBEL processes mentions sequentially within each document, inference can be parallelized across documents. In practice, we batch multiple documents and process them simultaneously. Each document maintains its own memory state, which is updated as mentions are resolved sequentially. When a document finishes, it is replaced by another document from the batch. Therefore, the sequential dependency limits parallelism within a document but does not prevent efficient batched inference over large collections.

\section{Effect of Semantic-Group Constraints}
\label{sec:semantic_group_ablation}

We compare sentence-level Baseline and LongBEL at both model sizes, with and without semantic-group constraints, which restrict decoding to the gold semantic group rather than the full candidate KB, as shown in Table~\ref{tab:semantic_group_ablation}.

\begin{table}[t]
\centering
\small
\begin{tabular}{lcc}
\toprule
\textbf{Dataset}
& \shortstack{\textbf{Without}\\\textbf{group}}
& \shortstack{\textbf{With}\\\textbf{group}} \\
\midrule
\multicolumn{3}{l}{\textit{Baseline-1B}} \\
MM-ST21pv & $69.1 {\scriptstyle \pm 1.0}$ & $73.1 {\scriptstyle \pm 1.0}$ \\
QUAERO-EMEA & $71.2 {\scriptstyle \pm 4.0}$ & $75.1 {\scriptstyle \pm 3.6}$ \\
SPACCC & $63.0 {\scriptstyle \pm 1.6}$ & $63.8 {\scriptstyle \pm 1.6}$ \\
\midrule
\multicolumn{3}{l}{\textit{Baseline-8B}} \\
MM-ST21pv & $70.9 {\scriptstyle \pm 0.9}$ & $75.0 {\scriptstyle \pm 0.9}$ \\
QUAERO-EMEA & $72.8 {\scriptstyle \pm 4.1}$ & $73.8 {\scriptstyle \pm 3.9}$ \\
SPACCC & $63.6 {\scriptstyle \pm 1.6}$ & $65.1 {\scriptstyle \pm 1.5}$ \\
\midrule
\multicolumn{3}{l}{\textit{LongBEL-1B}} \\
MM-ST21pv & $72.8 {\scriptstyle \pm 1.0}$ & $77.6 {\scriptstyle \pm 0.9}$ \\
QUAERO-EMEA & $74.1 {\scriptstyle \pm 3.4}$ & $74.5 {\scriptstyle \pm 3.7}$ \\
SPACCC & $61.9 {\scriptstyle \pm 1.5}$ & $63.1 {\scriptstyle \pm 1.6}$ \\
\midrule
\multicolumn{3}{l}{\textit{LongBEL-8B}} \\
MM-ST21pv & $75.1 {\scriptstyle \pm 0.9}$ & $79.3 {\scriptstyle \pm 0.8}$ \\
QUAERO-EMEA & $72.5 {\scriptstyle \pm 3.4}$ & $75.4 {\scriptstyle \pm 3.4}$ \\
SPACCC & $64.2 {\scriptstyle \pm 1.6}$ & $65.3 {\scriptstyle \pm 1.5}$ \\
\bottomrule
\end{tabular}
\caption{Recall@1 with and without gold semantic-group constraints. Values are percentages with half of the 95\% confidence interval.}
\label{tab:semantic_group_ablation}
\end{table}

Semantic-group constraints improve Recall@1 across all settings, with the largest gains on MM-ST21pv (+4.0–4.8 points; QUAERO-EMEA: +0.4–3.9; SPACCC: +0.8–1.5). Removing them lowers absolute performance but leaves LongBEL's advantage over the same-size baseline unchanged. As described in the benchmark setting, the same semantic-group constraints were used for LongBEL and all baselines that support them.

\end{document}